\documentclass[12pt]{article}

\usepackage{amsmath,amsthm, amsfonts, amssymb, amsxtra, amsopn}
\usepackage{pgfplots}
\pgfplotsset{compat=1.13}
\usepackage{pgfplotstable}
\usepackage{graphicx,grffile}
\usepackage{multirow}
\usepackage{booktabs}
\usepackage{listings}
\usepackage{cmap}
\usepackage{colortbl}
\usepackage{adjustbox}
\usepackage{epsfig}

\usepackage[tableposition=top,font=small,skip=5pt]{caption}
\usepackage{subcaption}
\usepackage{makecell} 

\usepackage[explicit]{titlesec}

\PassOptionsToPackage{hyphens}{url}

\usepackage{hyperref}
\hypersetup{colorlinks=true,linkcolor=black,citecolor=black,urlcolor=blue,filecolor=black}
\hypersetup{pdfpagemode=UseNone,pdfstartview=}

\definecolor{darkgreen}{rgb}{0.125,0.5,0.169}

\usepackage{enumitem}
\setlist[itemize]{noitemsep, topsep=0pt}

\advance\oddsidemargin by -0.35in
\advance\textwidth by 0.7in

\advance\topmargin by -0.45in
\advance\textheight by 0.9in

\long\def\symbolfootnotetext[#1]#2{\begingroup%
\def\thefootnote{\fnsymbol{footnote}}\footnotetext[#1]{#2}\endgroup}

\newcommand\dunderline[3][-1pt]{{%
  \sbox0{#3}%
  \ooalign{\copy0\cr\rule[\dimexpr#1-#2\relax]{\wd0}{#2}}}}
\def\uuu{\kern-1pt\dunderline{0.75pt}{\phantom{M}}}

\title{Transforming Chatbot Text: A Sequence-to-Sequence Approach} 

\author{Natesh Reddy\footnotemark[1]\ \ \ 
Mark Stamp\footnotemark[1]\,\,\footnotemark[2]}

\begin{document}

\symbolfootnotetext[1]{Department of Computer Science, San Jose State University}
\symbolfootnotetext[2]{mark.stamp$@$sjsu.edu}

\maketitle

\abstract
Due to advances in Large Language Models (LLMs) such as ChatGPT, 
the boundary between human-written text and AI-generated text 
has become blurred. Nevertheless, recent work has demonstrated that it is 
possible to reliably detect GPT-generated text.
In this paper, we adopt a novel strategy to adversarially transform 
GPT-generated text using sequence-to-sequence (Seq2Seq) 
models, with the goal of making the text more human-like. 
We experiment with the Seq2Seq models T5-small and BART
which serve to modify GPT-generated sentences to include  
linguistic, structural, and semantic components that may be more typical of 
human-authored text. Experiments show that classification models trained to 
distinguish GPT-generated text are significantly less accurate
when tested on text that has been modified by these Seq2Seq models. 
However, after retraining classification models on data
generated by our Seq2Seq technique, the models 
are able to distinguish the transformed GPT-generated 
text from human-generated text with high accuracy. This work adds to the 
accumulating knowledge of text transformation as a tool for both 
attack---in the sense of defeating classification models---and 
defense---in the sense of improved classifiers---thereby advancing 
our understanding of AI-generated text.

\section{Introduction}

Recent developments of Large Language Models (LLMs) such as 
GPT-4~\cite{openai2024gpt4technicalreport}, 
Claude~\cite{anthropic2023claudemodelcard}, and 
PaLM~\cite{chowdhery2022palmscalinglanguagemodeling}, 
can be viewed as major advances in the field of Natural Language Processing (NLP). 
The capabilities of machines to understand and generate output that is similar of the 
text written by humans has been revolutionized by these models. While this progress 
has numerous applications, it also brings a new set of challenges. 
For example, the task of distinguishing AI-generated text from human-written content has become 
more difficult. 
In addition, LLMs raise other concerns, including accountability, misuse, 
and fairness, highlighting the urgent need to strengthen detection 
capabilities~\cite{wu2025surveyllmdetection}. 

Researchers have developed classification systems to detect AI-generated content 
by analyzing linguistic features, structural patterns, and semantic 
cues~\cite{zhou2024humanizingmachinegeneratedcontentevading}. 
Although these systems perform well on clean datasets, they may be vulnerable to 
adversarial attacks. A recent study~\cite{DBLP:journals/corr/abs-2501-18998} 
revealed that small modifications to AI-generated text can significantly reduce
detection accuracy. This highlights a critical weakness in AI-generated text detection 
techniques used today.

In this research, we consider the use of a sequence-to-sequence (Seq2Seq) modeling approach, 
with the goal of transforming AI-generated text to a more human-like form. We fine tune popular
Seq2Seq models to modify GPT output, introducing more human-like characteristics, 
including more varied sentence structure, natural transitions, and increased linguistic diversity.

The research presented in this paper is focused on the following problems.
\begin{itemize}
\item First, we consider an adversarial attack, 
where we use Seq2Seq models to transform GPT-generated text, with the goal of 
making it more difficult to distinguish from human-generated text. 
We measure the success of this Seq2Seq approach by evaluating how well
models trained to distinguish GPT-generated text from human text perform when the
GPT-generated text is replaced by our Seq2Seq-transformed text.
\item We then explore ways to improve the robustness of classifiers against 
such adversarial attack by training classifiers on our Seq2Seq-transformed text. 
\end{itemize}
Our research provides a practical adversarial attack strategy, while also serving as a
data augmentation technique that can be used to improve classification accuracy.

The remainder of this paper is organized as follows. In Section~\ref{chap:related} we discuss relevant 
related work, with the focus on recent advances in the field of AI text detection and text
transformation techniques. Then in Section~\ref{chap:background}, we provide relevant background 
information on classification models, word embedding techniques, and the dataset used in 
our research. Section~\ref{chap:methodology} presents our proposed methodology using
Seq2Seq models, including the training process and evaluation metrics. In Section~\ref{chap:experiments} 
we present our experimental results. Finally, Section~\ref{chap:conclusion} provides a 
summary of our findings and we discuss future research directions.

\section{Related Work}\label{chap:related}

Continued improvements in Large Language Models (LLMs) 
has sparked interest in methods to distinguish AI-generated text from human writing. 
Early methods like~\cite{zellers2020defendingneuralfakenews}
mostly rely on comparing linguistic features and semantic patterns, 
while more recent approaches incorporate techniques such as word embeddings 
and transformer-based classifiers. 

Researchers have also implemented text transformation techniques to defeat classifiers. 
In this vein of research, Generative Adversarial Networks (GANs) are a powerful approach. 
However, in recent studies, Seq2Seq models such as T5~\cite{10.5555/3455716.3455856} 
and BART~\cite{lewis-etal-2020-bart} have demonstrated that such models can improve on GANs in
transformation tasks.

Generative Pre-Trained Transformers (GPT) can produce output that appears to be 
very similar to human writing. Such models have the ability to generate contextually 
appropriate text that is similar in structure, grammar, and fluency to human writing. 
However, there are subtle differences in stylistic patterns, word usage, complexity of sentences, 
and other structural features~\cite{zellers2020defendingneuralfakenews}. 
AI-generated text often lacks the natural flow, diversity in sentence structure, and emotional tone 
that are common in human-written content.  
These differences have been used to build detection 
systems that rely on semantic content analysis.

Word embeddings serve to convert words into a vector representation that captures both 
semantic meaning and syntactic structure. This has played a crucial role in enabling machine 
learning models to understand and process text, and word embeddings can serve as effective 
features for a wide variety of NLP tasks~\cite{turian-etal-2010-word}. Embeddings are especially 
valued for their ability to encode surprisingly accurate syntactic and semantic relationships between words~\cite{mikolov2013efficientestimationwordrepresentations}.

There are various embedding models, each with their unique strengths. For instance, 
Word2Vec~\cite{mikolov2013efficientestimationwordrepresentations}  
determines word relationships by using context-focused methods, including 
Continuous Bag-of-Words (CBOW) and Skip-Gram. In contrast, GloVe~\cite{pennington-etal-2014-glove} 
derives embeddings based
on how often words appear together, thus allowing for a better understanding of context. 
Bidirectional Encoder Representations from Transformers (BERT)~\cite{devlin-etal-2019-bert} 
revolutionized the field by 
considering bidirectional context. BERT identifies language details missed by earlier methods, 
including Word2Vec and GloVe. This allows BERT to better understand linguistic cues and
semantic meaning, leading to major improvements in a wide range of NLP applications.

It has been previously demonstrated in studies such as~\cite{zellers2020defendingneuralfakenews} 
and~\cite{wallace-etal-2019-universal}, that embeddings can significantly improve classification accuracy, 
especially when paired with learning architectures such as Multilayer Perceptrons (MLPs) 
or Long Short-Term Memory (LSTM) networks. 
These approaches are known to perform well, 
however, they are sometimes vulnerable to adversarial attacks~\cite{ebrahimi-etal-2018-hotflip}.

Seq2Seq models have become a foundational technique 
in NLP. Examples of the use of such models include 
machine translation, text summarization, and style transfer. Seq2Seq architectures are
effective at handling variable-length input and output, which makes them suitable for
a wide range of transformation tasks. The work in~\cite{sutskever2014sequencesequencelearningneural} 
demonstrated the capability of Seq2Seq models
by surpassing traditional phrase-based statistical machine 
translation systems. The effectiveness of Seq2Seq models for transforming AI-generated text 
into more human like text is a primary focus of the research in this paper. 

Data augmentation is a common strategy to increase the performance and generalization of 
text classification models. Traditional augmentation methods try to introduce diversity to 
textual data without altering the core meaning. This can be done by synonym substitution, 
for example. These methods can also help prevent overfitting and make models more robust. 
In the field of text generation, style transfer mechanisms can be used to create 
new, augmentation samples~\cite{chai2025textdataaugmentationlarge}.

The challenging task of detecting AI-generated text is relevant across multiple 
domains~\cite{gozalobrizuela2023surveygenerativeaiapplications}.
Here, we highlight a few such domains.

\begin{description}
\item[Content Moderation and Misinformation Detection]\!--- Since the amount of AI-generated 
data is increasing daily, to prevent the spread of misinformation 
and protect the integrity of social media platforms, it is important to be able to detect 
and flag AI-generated content~\cite{GonganeDetrimentalContent2022}.
\item[Education and Academic Integrity]\!--- Detections systems 
that are more robust are needed to protect academic integrity.
For example, AI writing tools are capable of completing assignments 
with minimal input from a student, and hence academic institutions need better methods to identify 
and flag AI-generated submissions~\cite{herbold2023aiwriteessayme}. 
\item[Authorship Attribution]\!--- Determining the authorship of content is critical in many
scenarios including, for example, legal cases, intellectual property disputes, 
and online investigations~\cite{masood2025intellectual}. Tools that can differentiate between 
AI-generated and human-generated content can aid the task of authorship verification.
\item[Content Personalization]\!--- Businesses have started relying on chatbots and 
automated writing tools for the purpose of improving user engagement.
More human-like AI-generated content would be useful in this context~\cite{RUNGRUANGJIT2024100523}.
\item[Cybersecurity]\!--- Content that is generated by an LLM can makes the task of detecting
phishing email and related scams more challenging, since such scams can be customized for
each user~\cite{10466545}. Therefore, robust detection systems need to be able to identify AI-generated content.
\item[Journalism and Media]\!--- Some media outlets have begun experimenting with AI-written articles. 
AI detection systems can play a vital role in ensuring that readers know the true origin of 
content~\cite{Kreps_McCain_Brundage_2022}.
\end{description}

In summary, the vast improvements in the quality of AI-generated text provided by LLMs
bring both benefits and new challenges. In this research, we focus on the challenge of detecting
AI-generated text. In the next section, we turn our attention to the techniques used in our 
experiments, and in subsequent sections we discuss our methodology in detail and
provide our experimental results.

\section{Background}\label{chap:background}

This section serves as an overview of the various learning techniques used in our experiments. 
We start by reviewing the classification models that we use. 
Then we look at the word embedding techniques that we consider, 
which are essential for transforming textual data into
a format that is more useful for machine learning models. 
We also introduce the Sequence-to-Sequence models that we use
to transform AI text into a more human-like form..
Lastly, we discuss the dataset 
that we use in our experiments.

\subsection{Classification Models}

For classification of text, we consider both classic machine learning (ML) and deep learning (DL) models. 
The classic ML techniques that we consider are Logistic Regression, Random Forest, and XGBoost,
while the DL techniques are Multilayer Perceptron, a Deep Neural Network, 
and Long Short-Term Memory networks. 

\subsubsection{Logistic Regression}

Linear Regression (LR) is computationally inexpensive, the results are interpretable, and 
the technique often performs surprisingly well in practice.
LR models the probability that the input belongs to a particular class by applying 
the sigmoid function to a linear combination of input features. Although it is an extremely simple model, 
LR performs well when the data is linearly separable, and it is often used to establish a baseline
for more advanced learning models~\cite{boateng2019logisticregression}. 

\subsubsection{Random Forest}

Random Forest (RF) is an ensemble-based learning technique based on decision trees. 
While decision trees are simple, they tend to overfit the data. Random Forests avoid
this overfitting issue by training a large number of decision trees on subsets of the
data and features---a process known as bagging. Random Forests are known for their 
robustness, generalization capabilities, their ability to handle noisy and 
high-dimensional data, and they often achieve results that are competitive with 
costly DL techniques~\cite{stamp2022introduction}.

\subsubsection{XGBoost}

Boosting is a general ensemble technique that combines relatively weak models into a single, 
much stronger model. Extreme Gradient Boosting (XGB) builds an ensemble of decision trees 
sequentially---each new tree that is added is selected so as to maximize the improvement in the
overall model. XGB supports efficient parallel computation and reduces overfitting by 
the use of regularization, and it has become one of the most widely used models 
in real-world ML applications, due to its performance and efficiency~\cite{Chen_2016}.

\subsubsection{Multilayer Perceptron}

Multilayer Perceptron (MLP) is a fully connected feedforward artificial neural network. 
An MLP includes an input layer, an output layer, and one or more hidden layers. 
Nonlinear activation function are applied to each layer which enables the network 
to learn complex patterns. MLP is a fundamental deep learning approach that
performs well for many tasks, although MLPs are not designed to effectively deal with sequential 
dependencies~\cite{yu2023multilayerperceptrontrainabilityexplained}.

\subsubsection{Deep Neural Network}

We use the term Deep Neural Network (DNN) to refer to an MLP architecture 
with multiple hidden layers. As compared to a (shallow) MLP, a DNN is more capable 
of learning complex patterns, but DNNs require more training data, and they are more
costly to train. DNNs have been widely used in various NLP tasks.

\subsubsection{Long Short Term Memory Models}

Long Short Term Memory (LSTM) models were introduced to handle vanishing and exploding
gradients, which are inherent problems in generic 
Recurrent Neural Networks (RNN). An LSTM architecture is a 
highly specialized form of RNN that includes a complex gating structure. 
These gates serve to regulate updates to and from long-term memory, thereby mitigating the 
aforementioned gradient issues. LSTMs are a strong choice for a variety of NLP tasks, 
since they can effectively account for sequential 
dependencies~\cite{vennerd2021longshorttermmemoryrnn}.

\subsection{Word Embeddings}

As mentioned above, word embeddings are a core component in most NLP systems. 
The purpose of word embeddings is to preserve semantic and syntactic meaning
when converting textual data into numerical vectors. Machine learning models 
generally perform much more effectively when embeddings are used. In this context,
word embeddings can be viewed as a feature engineering step that serves to
provide improved data to the downstream learning model.
This section provides a brief overview of the three word embedding techniques
that we employ, namely, Word2Vec, GloVe, and BERT.

\subsubsection{Word2Vec}

Word2Vec trains a simple neural network model on text, and then uses the weights of the model
as vector representations of the input words. 
Words that are closer together (in terms of cosine similarity) in the resulting vector representation 
are semantically more closely related. Continuous Bag-of-Words (CBOW)
and Skip-Gram approaches are supported in Word2Vec. The Skip-Gram technique uses a target 
word to predict its context, while CBOW predicts the target word based on the surrounding context. 
We use CBOW for our Word2Vec embeddings, since it tends to perform better on larger 
datasets~\cite{mikolov2013efficientestimationwordrepresentations}.

\subsubsection{GloVe}

Global Vectors for Word Representation (GloVe) is another popular method for generating word embeddings. 
GloVe builds a word-context co-occurrence matrix from a large corpus and learns word vectors by 
factorizing the resulting matrix. This approach captures 
global statistics related to word usage by minimizing the difference between co-occurrence 
probabilities and the dot product of corresponding word vectors. The generated embeddings 
are rich in both semantic and syntactic information, the vector are dense, 
and GloVe vectors encode meaningful relationships into the embedding 
space~\cite{pennington-etal-2014-glove}.

\subsubsection{BERT}

Bidirectional Encoder Representations from Transformers (BERT) provides contextual embeddings, 
that is, the same word can have different representation depending on the surrounding context. 
Note that this is in contrast to Word2Vec and GloVe, both of which assign a single vector per word.
As the name suggests, BERT processes text bidirectionally, using transformer encoders. 
The use of Masked Language Modeling (MLM) 
and Next Sentence Prediction (NSP) techniques, enable BERT to learn deeper linguistic patterns. 
BERT embeddings have enabled models to understand complex sentence structures, and
hence this technique is responsible for significant advances in NLP~\cite{devlin-etal-2019-bert}. 

\subsubsection{T5-Small}

Text-to-Text Transfer Transformer (T5)
is an encoder-decoder model. T5
has achieved state of the art results across a diverse range NLP tasks,
including translation, summarization, and text classification~\cite{10.5555/3455716.3455856}.
We use the T5-small model, due to its balance between performance and computational efficiency. 
By fine-tuning the pretrained T5-small model, we aim to use it to transform our GPT-generated text 
into a style that is more similar to that of human written content. Examples of our T5-transformed text
appear in the Appendix.

\subsubsection{BART}

Bidirectional and Auto-Regressive Transformers (BART) is a denoising autoencoder, 
which was introduced for training sequence to sequence models. 
BART is particularly efficient for text generation task, as it combines the bidirectional encoder of BERT 
with autoregressive decoder of GPT. In BART, an arbitrary noise function is used to modify the text, 
which is then used to train BART to generate the original text. This method allows the model to learn 
complex dependencies in text~\cite{lewis-etal-2020-bart}. 

We fine-tune a pretrained BART model. As with T5-small, our purpose for using BART is to 
generate more fluent, coherent, and human-like text from a dataset of AI-generated text.
Examples of our BART-transformed text appear in the Appendix.

\subsection{Dataset}\label{section:dataset}

The dataset used in this work was introduced in~\cite{godghase2024distinguishingchatbothuman},
where it was created for the task of distinguishing between human-written and chatbot-generated content. 
The dataset is an ideal fit for the research presented in this paper, since it includes a large 
amount of human written content and corresponding GPT-generated text.

The original dataset includes two classes, namely, human-generated text and GPT-generated text. 
The human-written samples were sourced from the publicly available WikiHow 
dataset~\cite{koupaee2018wikihowlargescaletext}. Each GPT-generated text 
was created using a structured prompt, based on a human-written sample.
Table~\ref{tab:19} summarizes the key statistics of the dataset. 
We observe that GPT-generated paragraphs are typically longer and more verbose in 
comparison to human written text. Occasionally, the GPT model also produced multi-paragraph 
output for a single prompt, contributing to a slightly higher total paragraph count in the GPT class. 
See~\cite{godghase2024distinguishingchatbothuman} for additional details on this dataset.

\begin{table}[!htb]
\centering
\caption{Dataset details}\label{tab:19}
\adjustbox{scale=0.85}{
\begin{tabular}{c|cccc}\toprule
\multirow{2}{*}{\textbf{Class}} & \multirow{2}{*}{\textbf{Paragraphs}} & \multirow{2}{*}{\textbf{Words}} 
	& \multirow{2}{*}{\textbf{Characters}} & \textbf{Average words}\\
    	&&&&  \textbf{per paragraph} \\ \midrule
Human & 784,636 & 54,005,604 & 307,005,548 & 68.83 \\ 
ChatGPT & 920,259 & 75,474,378 & 474,396,685 & 82.01 \\ 
\bottomrule
\end{tabular}
}
\end{table}

\section{Methodology}\label{chap:methodology}

In an effort to make GPT-generated text more similar in tone and structure 
to human-written content, we use a Seq2Seq-based transformation pipeline. 
As opposed to relying on adversarial training with GANs, our method relies 
on the powerful transformer-based encoder-decoder 
models T5-small and BART to perform controlled style transfer. 
The goal is to preserve the semantic meaning of 
the input, while transforming surface characteristics to more closely resemble human-authored writing. 

Our Seq2Seq models are fine-tuned in a supervised manner using paired datasets. 
Each input is a GPT-generated sentence prefixed with a task-specific instruction 
(i.e., ``humanize'') and the corresponding target is the human-written 
version which was taken from the same semantic context. 
This formulation allows us to take advantage of the text-to-text
transformation capabilities of T5 and BART.

We use an~80:20 split for training and evaluation. 
All experiments are based on balanced data consisting of~10,000 samples
(i.e., 5,000 samples from each class),
which implies that accuracies are based on~2,000 samples.

Text is tokenized using model-specific tokenizers. 
Once tokenized, the token sequence is then padded or truncated to a specified maximum sequence length. 
After analyzing the distribution of the text in our dataset, as shown in Table~\ref{tab:length-stats}, 
we chose the maximum sequence length to be~512
to align with the model positional encoding limit and to maintain 
consistency with pretrained configurations. Both 
T5-small and BART tend to overfit when trained on short sequences, so we chose
the length to capture the widest possible range of sequence lengths.
The input and target sequences both are encoded into token IDs, and attention 
masks are applied before being passed to the model. The attention masks are 
automatically generated during tokenization to differentiate between tokens and 
padded positions. Each token represents a sub-word unit derived from the tokenizers vocabulary. 
Depending on the tokenizer (e.g., T5 or BART), a single token may correspond to a whole word, 
a part of a word, or punctuation. These sub-word level tokenizations enable the model to handle 
rare words and morphological variations more effectively.

\begin{table}[!htb]
\centering
\caption{Distribution of text in dataset}\label{tab:length-stats}
\adjustbox{scale=0.85}{
\begin{tabular}{c|ccc}\toprule
\multirow{2}{*}{\textbf{Dataset}} & \multicolumn{3}{c}{\textbf{Sequence length}} \\ 
 & \textbf{Minimum} & \textbf{Maximum} & \textbf{Average} \\ \midrule
Human-generated & 16 & 478 & 129.31  \\ 
GPT-generated & 21 & 252 & 135.99 \\ 
\bottomrule
\end{tabular}
}
\end{table}


To prevent overfitting, we also use AdamW (Adam with Weight 
decay) as our optimizer~\cite{loshchilov2019decoupledweightdecayregularization}. A linear learning 
rate scheduler with warm-up steps is also applied to improve convergence. This helps causes
the learning rate to gradually increases during the early steps before decaying linearly. To reduce 
memory usage and training time, we enabled 16-bit floating point (FP16) precision during training. 
This lower-precision representation of weights significantly reduces GPU memory consumption 
while maintaining comparable model performance, and is especially effective for 
transformer-based architectures like T5 and BART.

We train each model up to a maximum of~20 epochs, saving a checkpoint at each epoch. 
We apply early stopping based on the model loss to reduce needless computation and
to avoid overfitting. Additionally, we apply label smoothing, a regularization technique that 
prevents the model from becoming overly confident in its predictions. Instead of assigning a 
probability of~1 to the correct class and~0 to all others, label smoothing distributes a small portion 
of the probability to incorrect classes. This improve generalization and mitigates overfitting.

Cross-entropy between the predicted token probabilities 
and target sequence tokens is used as the loss function. Since the output is a sequence of tokens, 
the model learns to generate each word step-by-step, conditioned on the previous output and the 
encoded representation of the input. 

During inference, we use decoding strategies of beam search and top-$k$/top-$p$ sampling with repetition penalty to 
generate the humanized version of GPT text. These strategies control how the model selects the next word in a sequence, 
balancing fluency and diversity in the generated text. The output is then passed through 
a post-processing step to remove padding sequences and to decode the token IDs back to 
natural language.

Using our trained T5-small and BART Seq2Seq models, two new datasets are created, 
which we refer to as the T5-generated dataset and the BART-generated dataset.
Note that this gives us a total of four distinct datasets---in addition to the T5-generated and
BART-generated datasets, we have a human-generated 
dataset consisting of WikiHow paragraphs, and the GPT-generated dataset 
from~\cite{godghase2024distinguishingchatbothuman} 
that was derived from the human-generated data. Note also that
both the T5-generated and BART-generated datasets are post-processed
versions of the GPT-generated dataset.

For all of our classification experiments, we test each of the embedding techniques discussed in 
Section~\ref{chap:background}, namely, Word2Vec, GloVe, and BERT. 
For each embedding technique, we experiment 
with all of the classifiers discussed in Section~\ref{chap:background}, namely,
LR, RF, XGB, MLP, DNN, and LSTM.
We use accuracy to evaluate the results of all of our experiments.

\section{Experiments and Results}\label{chap:experiments}

In this section, we present our experimental results. We conduct extensive experiments
for each of the following cases.
\begin{description}
\item[Baseline]\!---
For our baseline case, we train and test classification models to distinguish 
human-generated text from GPT-generated text.
\item[Baseline Models Tested on Transformed Text]\!---
As a second set of experiments, we 
test the classification models derived in the baseline case to determine how well they can
distinguish human-generated text from our T5-generated data
and our BART-generated data. 
This case determines how well these Seq2Seq
transformation techniques perform, with respect to defeating classifiers 
that have been trained to distinguish human-generated
from GPT-generated text.
\item[Models Retrained on Transformed Text]\!---
As a final case, 
we train and test classification models to distinguish 
human-generated text from our Seq2Seq-generated text. 
This case provides insight into the nature of our T5-generated 
and BART-generated data. Specifically, this case is designed to 
determine how successful our T5-small and BART models are in ``humanizing''
the GPT-generated text.
\end{description}
In the first case, we train six learning models on three different
embeddings for a total of~18 experiments. In both of the latter two cases, 
these same combinations are tested with both T5-small and BART transformations, 
for a total of~36 experimental in each case. Thus, we have
a total of~90 distinct experiments.

\subsection{Baseline Results}\label{sect:case1}

In this case, we train and test on the human-generated and GPT-generated text,
using each of Word2Vec, GloVe, and BERT embeddings for each of the LR, 
RF, XGB, MLP, DNN, and LSTM models.
The accuracies obtained for all of these experiments 
are given in Table~\ref{tab:baseline-all}. 
We observe that BERT embeddings for the DNN model perform best,
while DNN is consistently best for each embedding technique, while MLP
performs nearly as well as DNN.

\begin{table}[!htb]
\centering\renewcommand{\arraystretch}{0.95}
\caption{Baseline accuracies}\label{tab:baseline-all}
\adjustbox{scale=0.8}{
\begin{tabular}{c|cc}\toprule
\textbf{Embedding} & \textbf{Model} & \textbf{Accuracy}\\ \midrule
\multirow{6}{*}{Word2Vec} 
& LR & 0.9300 \\ 
& RF & 0.9205 \\ 
& XGB & 0.9345 \\ 
& MLP & 0.9395 \\ 
& DNN & \textbf{0.9490} \\ 
& LSTM & 0.9315 \\  \midrule
\multirow{6}{*}{GloVe} 
& LR & 0.9510 \\ 
& RF & 0.9535 \\ 
& XGB & 0.9600 \\ 
& MLP & 0.9730 \\ 
& DNN & \textbf{0.9740} \\ 
& LSTM & 0.9605 \\  \midrule
\multirow{6}{*}{BERT}
& LR & 0.9825 \\ 
& RF & 0.9585 \\ 
& XGB & 0.9720 \\ 
& MLP & 0.9820 \\ 
& DNN & \textbf{0.9840} \\ 
& LSTM & 0.8495 \\  \bottomrule
\end{tabular}
}
\end{table}

\begin{figure}[!htb]
\advance\tabcolsep by-6.5pt
\centering
\begin{tabular}{ccc}
\begin{adjustbox}{scale=0.875}
\begin{tikzpicture}[scale=0.5]
    \begin{axis}[
        width=7.5cm,
        height=7.5cm,
	colormap={bluewhite}{color=(white) rgb255=(100,149,237)},
        xticklabels={
        \texttt{GPT},
        \texttt{Human},
        },
        xtick={0,...,1},
        xtick style={draw=none},
	xticklabel style={anchor=east,rotate=60,yshift=-5pt,font=\tt,scale=1.5},
        yticklabels={
        \texttt{GPT},
        \texttt{Human},
        },
        ytick={0,...,1},
        ytick style={draw=none},
        enlargelimits=false,
        yticklabel style={xshift=2.5pt,font=\tt,scale=1.5},
        colorbar,
        colorbar style={
            ytick={0.00,0.20,0.40,0.60,0.80,1.00},
            yticklabels={0.00,0.20,0.40,0.60,0.80,1.00},
            yticklabel={\pgfmathprintnumber\tick},
            yticklabel style={font=\tt, scale=1.5,
            		/pgf/number format/fixed,
			/pgf/number format/fixed zerofill,
			/pgf/number format/precision=2}
        },
        point meta min=0.0,
        point meta max=1.0,
        nodes near coords={\pgfmathprintnumber\pgfplotspointmeta},
        nodes near coords black white/.style={
            small value/.style={
                yshift=-10pt,
                text=black,
                /pgf/number format/fixed,
                /pgf/number format/precision=3,
                /pgf/number format/zerofill=true,
                scale=1.5,
            },
            large value/.style={
                yshift=-10pt,
                text=white,
                /pgf/number format/fixed,
                /pgf/number format/precision=3,
                /pgf/number format/zerofill=true,
                scale=1.5,
            },
            every node near coord/.style={
                check for zero/.code={
                    \pgfmathfloatifflags{\pgfplotspointmeta}{0}{
                        \pgfkeys{/tikz/coordinate}
                    }{
                        \begingroup
                        \pgfkeys{/pgf/fpu}
                        \pgfmathparse{\pgfplotspointmeta<#1}
                        \global\let\result=\pgfmathresult
                        \endgroup
                        %
                        %
                        \pgfmathfloatcreate{1}{1.0}{0}
                        \let\ONE=\pgfmathresult
                        \ifx\result\ONE
                            \pgfkeysalso{/pgfplots/small value}
                        \else
                            \pgfkeysalso{/pgfplots/large value}
                        \fi
                    }
                },
                check for zero,
            },
        },
        nodes near coords black white=0.5,
    ]
        \addplot[
            matrix plot,
            mesh/cols=2,
            point meta=explicit,draw=gray
        ] table [meta=C] {
            x y C
0 0 0.957
1 0 0.043
0 1 0.055
1 1 0.945
         };
    \end{axis}
\end{tikzpicture}
\end{adjustbox}
&
\begin{adjustbox}{scale=0.875}
\begin{tikzpicture}[scale=0.5]
    \begin{axis}[
        width=7.5cm,
        height=7.5cm,
	colormap={bluewhite}{color=(white) rgb255=(100,149,237)},
        xticklabels={
        \texttt{GPT},
        \texttt{Human},
        },
        xtick={0,...,1},
        xtick style={draw=none},
	xticklabel style={anchor=east,rotate=60,yshift=-5pt,font=\tt,scale=1.5},
        yticklabels={
        \texttt{GPT},
        \texttt{Human},
        },
        ytick={0,...,1},
        ytick style={draw=none},
        enlargelimits=false,
        yticklabel style={xshift=2.5pt,font=\tt,scale=1.5},
        colorbar,
        colorbar style={
            ytick={0.00,0.20,0.40,0.60,0.80,1.00},
            yticklabels={0.00,0.20,0.40,0.60,0.80,1.00},
            yticklabel={\pgfmathprintnumber\tick},
            yticklabel style={font=\tt, scale=1.5,
            		/pgf/number format/fixed,
			/pgf/number format/fixed zerofill,
			/pgf/number format/precision=2}
        },
        point meta min=0.0,
        point meta max=1.0,
        nodes near coords={\pgfmathprintnumber\pgfplotspointmeta},
        nodes near coords black white/.style={
            small value/.style={
                yshift=-10pt,
                text=black,
                /pgf/number format/fixed,
                /pgf/number format/precision=3,
                /pgf/number format/zerofill=true,
                scale=1.5,
            },
            large value/.style={
                yshift=-10pt,
                text=white,
                /pgf/number format/fixed,
                /pgf/number format/precision=3,
                /pgf/number format/zerofill=true,
                scale=1.5,
            },
            every node near coord/.style={
                check for zero/.code={
                    \pgfmathfloatifflags{\pgfplotspointmeta}{0}{
                        \pgfkeys{/tikz/coordinate}
                    }{
                        \begingroup
                        \pgfkeys{/pgf/fpu}
                        \pgfmathparse{\pgfplotspointmeta<#1}
                        \global\let\result=\pgfmathresult
                        \endgroup
                        %
                        %
                        \pgfmathfloatcreate{1}{1.0}{0}
                        \let\ONE=\pgfmathresult
                        \ifx\result\ONE
                            \pgfkeysalso{/pgfplots/small value}
                        \else
                            \pgfkeysalso{/pgfplots/large value}
                        \fi
                    }
                },
                check for zero,
            },
        },
        nodes near coords black white=0.5,
    ]
        \addplot[
            matrix plot,
            mesh/cols=2,
            point meta=explicit,draw=gray
        ] table [meta=C] {
            x y C
0 0 0.957
1 0 0.043
0 1 0.050
1 1 0.950
         };
    \end{axis}
\end{tikzpicture}
\end{adjustbox}
&
\begin{adjustbox}{scale=0.875}
\begin{tikzpicture}[scale=0.5]
    \begin{axis}[
        width=7.5cm,
        height=7.5cm,
	colormap={bluewhite}{color=(white) rgb255=(100,149,237)},
        xticklabels={
        \texttt{GPT},
        \texttt{Human},
        },
        xtick={0,...,1},
        xtick style={draw=none},
	xticklabel style={anchor=east,rotate=60,yshift=-5pt,font=\tt,scale=1.5},
        yticklabels={
        \texttt{GPT},
        \texttt{Human},
        },
        ytick={0,...,1},
        ytick style={draw=none},
        enlargelimits=false,
        yticklabel style={xshift=2.5pt,font=\tt,scale=1.5},
        colorbar,
        colorbar style={
            ytick={0.00,0.20,0.40,0.60,0.80,1.00},
            yticklabels={0.00,0.20,0.40,0.60,0.80,1.00},
            yticklabel={\pgfmathprintnumber\tick},
            yticklabel style={font=\tt, scale=1.5,
            		/pgf/number format/fixed,
			/pgf/number format/fixed zerofill,
			/pgf/number format/precision=2}
        },
        point meta min=0.0,
        point meta max=1.0,
        nodes near coords={\pgfmathprintnumber\pgfplotspointmeta},
        nodes near coords black white/.style={
            small value/.style={
                yshift=-10pt,
                text=black,
                /pgf/number format/fixed,
                /pgf/number format/precision=3,
                /pgf/number format/zerofill=true,
                scale=1.5,
            },
            large value/.style={
                yshift=-10pt,
                text=white,
                /pgf/number format/fixed,
                /pgf/number format/precision=3,
                /pgf/number format/zerofill=true,
                scale=1.5,
            },
            every node near coord/.style={
                check for zero/.code={
                    \pgfmathfloatifflags{\pgfplotspointmeta}{0}{
                        \pgfkeys{/tikz/coordinate}
                    }{
                        \begingroup
                        \pgfkeys{/pgf/fpu}
                        \pgfmathparse{\pgfplotspointmeta<#1}
                        \global\let\result=\pgfmathresult
                        \endgroup
                        %
                        %
                        \pgfmathfloatcreate{1}{1.0}{0}
                        \let\ONE=\pgfmathresult
                        \ifx\result\ONE
                            \pgfkeysalso{/pgfplots/small value}
                        \else
                            \pgfkeysalso{/pgfplots/large value}
                        \fi
                    }
                },
                check for zero,
            },
        },
        nodes near coords black white=0.5,
    ]
        \addplot[
            matrix plot,
            mesh/cols=2,
            point meta=explicit,draw=gray
        ] table [meta=C] {
            x y C
0 0 0.962
1 0 0.038
0 1 0.042
1 1 0.958
         };
    \end{axis}
\end{tikzpicture}
\end{adjustbox}
\\[-0.5ex]
\adjustbox{scale=0.9}{(a) LR}
&
\adjustbox{scale=0.9}{(b) RF}
&
\adjustbox{scale=0.9}{(c) XGB}
\\ \\[-2ex]
\begin{adjustbox}{scale=0.875}
\begin{tikzpicture}[scale=0.5]
    \begin{axis}[
        width=7.5cm,
        height=7.5cm,
	colormap={bluewhite}{color=(white) rgb255=(100,149,237)},
        xticklabels={
        \texttt{GPT},
        \texttt{Human},
        },
        xtick={0,...,1},
        xtick style={draw=none},
	xticklabel style={anchor=east,rotate=60,yshift=-5pt,font=\tt,scale=1.5},
        yticklabels={
        \texttt{GPT},
        \texttt{Human},
        },
        ytick={0,...,1},
        ytick style={draw=none},
        enlargelimits=false,
        yticklabel style={xshift=2.5pt,font=\tt,scale=1.5},
        colorbar,
        colorbar style={
            ytick={0.00,0.20,0.40,0.60,0.80,1.00},
            yticklabels={0.00,0.20,0.40,0.60,0.80,1.00},
            yticklabel={\pgfmathprintnumber\tick},
            yticklabel style={font=\tt, scale=1.5,
            		/pgf/number format/fixed,
			/pgf/number format/fixed zerofill,
			/pgf/number format/precision=2}
        },
        point meta min=0.0,
        point meta max=1.0,
        nodes near coords={\pgfmathprintnumber\pgfplotspointmeta},
        nodes near coords black white/.style={
            small value/.style={
                yshift=-10pt,
                text=black,
                /pgf/number format/fixed,
                /pgf/number format/precision=3,
                /pgf/number format/zerofill=true,
                scale=1.5,
            },
            large value/.style={
                yshift=-10pt,
                text=white,
                /pgf/number format/fixed,
                /pgf/number format/precision=3,
                /pgf/number format/zerofill=true,
                scale=1.5,
            },
            every node near coord/.style={
                check for zero/.code={
                    \pgfmathfloatifflags{\pgfplotspointmeta}{0}{
                        \pgfkeys{/tikz/coordinate}
                    }{
                        \begingroup
                        \pgfkeys{/pgf/fpu}
                        \pgfmathparse{\pgfplotspointmeta<#1}
                        \global\let\result=\pgfmathresult
                        \endgroup
                        %
                        %
                        \pgfmathfloatcreate{1}{1.0}{0}
                        \let\ONE=\pgfmathresult
                        \ifx\result\ONE
                            \pgfkeysalso{/pgfplots/small value}
                        \else
                            \pgfkeysalso{/pgfplots/large value}
                        \fi
                    }
                },
                check for zero,
            },
        },
        nodes near coords black white=0.5,
    ]
        \addplot[
            matrix plot,
            mesh/cols=2,
            point meta=explicit,draw=gray
        ] table [meta=C] {
            x y C
0 0 0.972
1 0 0.028
0 1 0.026
1 1 0.974
         };
    \end{axis}
\end{tikzpicture}
\end{adjustbox}
&
\begin{adjustbox}{scale=0.875}
\begin{tikzpicture}[scale=0.5]
    \begin{axis}[
        width=7.5cm,
        height=7.5cm,
	colormap={bluewhite}{color=(white) rgb255=(100,149,237)},
        xticklabels={
        \texttt{GPT},
        \texttt{Human},
        },
        xtick={0,...,1},
        xtick style={draw=none},
	xticklabel style={anchor=east,rotate=60,yshift=-5pt,font=\tt,scale=1.5},
        yticklabels={
        \texttt{GPT},
        \texttt{Human},
        },
        ytick={0,...,1},
        ytick style={draw=none},
        enlargelimits=false,
        yticklabel style={xshift=2.5pt,font=\tt,scale=1.5},
        colorbar,
        colorbar style={
            ytick={0.00,0.20,0.40,0.60,0.80,1.00},
            yticklabels={0.00,0.20,0.40,0.60,0.80,1.00},
            yticklabel={\pgfmathprintnumber\tick},
            yticklabel style={font=\tt, scale=1.5,
            		/pgf/number format/fixed,
			/pgf/number format/fixed zerofill,
			/pgf/number format/precision=2}
        },
        point meta min=0.0,
        point meta max=1.0,
        nodes near coords={\pgfmathprintnumber\pgfplotspointmeta},
        nodes near coords black white/.style={
            small value/.style={
                yshift=-10pt,
                text=black,
                /pgf/number format/fixed,
                /pgf/number format/precision=3,
                /pgf/number format/zerofill=true,
                scale=1.5,
            },
            large value/.style={
                yshift=-10pt,
                text=white,
                /pgf/number format/fixed,
                /pgf/number format/precision=3,
                /pgf/number format/zerofill=true,
                scale=1.5,
            },
            every node near coord/.style={
                check for zero/.code={
                    \pgfmathfloatifflags{\pgfplotspointmeta}{0}{
                        \pgfkeys{/tikz/coordinate}
                    }{
                        \begingroup
                        \pgfkeys{/pgf/fpu}
                        \pgfmathparse{\pgfplotspointmeta<#1}
                        \global\let\result=\pgfmathresult
                        \endgroup
                        %
                        %
                        \pgfmathfloatcreate{1}{1.0}{0}
                        \let\ONE=\pgfmathresult
                        \ifx\result\ONE
                            \pgfkeysalso{/pgfplots/small value}
                        \else
                            \pgfkeysalso{/pgfplots/large value}
                        \fi
                    }
                },
                check for zero,
            },
        },
        nodes near coords black white=0.5,
    ]
        \addplot[
            matrix plot,
            mesh/cols=2,
            point meta=explicit,draw=gray
        ] table [meta=C] {
            x y C
0 0 0.972
1 0 0.028
0 1 0.024
1 1 0.976
         };
    \end{axis}
\end{tikzpicture}
\end{adjustbox}
&
\begin{adjustbox}{scale=0.875}
\begin{tikzpicture}[scale=0.5]
    \begin{axis}[
        width=7.5cm,
        height=7.5cm,
	colormap={bluewhite}{color=(white) rgb255=(100,149,237)},
        xticklabels={
        \texttt{GPT},
        \texttt{Human},
        },
        xtick={0,...,1},
        xtick style={draw=none},
	xticklabel style={anchor=east,rotate=60,yshift=-5pt,font=\tt,scale=1.5},
        yticklabels={
        \texttt{GPT},
        \texttt{Human},
        },
        ytick={0,...,1},
        ytick style={draw=none},
        enlargelimits=false,
        yticklabel style={xshift=2.5pt,font=\tt,scale=1.5},
        colorbar,
        colorbar style={
            ytick={0.00,0.20,0.40,0.60,0.80,1.00},
            yticklabels={0.00,0.20,0.40,0.60,0.80,1.00},
            yticklabel={\pgfmathprintnumber\tick},
            yticklabel style={font=\tt, scale=1.5,
            		/pgf/number format/fixed,
			/pgf/number format/fixed zerofill,
			/pgf/number format/precision=2}
        },
        point meta min=0.0,
        point meta max=1.0,
        nodes near coords={\pgfmathprintnumber\pgfplotspointmeta},
        nodes near coords black white/.style={
            small value/.style={
                yshift=-10pt,
                text=black,
                /pgf/number format/fixed,
                /pgf/number format/precision=3,
                /pgf/number format/zerofill=true,
                scale=1.5,
            },
            large value/.style={
                yshift=-10pt,
                text=white,
                /pgf/number format/fixed,
                /pgf/number format/precision=3,
                /pgf/number format/zerofill=true,
                scale=1.5,
            },
            every node near coord/.style={
                check for zero/.code={
                    \pgfmathfloatifflags{\pgfplotspointmeta}{0}{
                        \pgfkeys{/tikz/coordinate}
                    }{
                        \begingroup
                        \pgfkeys{/pgf/fpu}
                        \pgfmathparse{\pgfplotspointmeta<#1}
                        \global\let\result=\pgfmathresult
                        \endgroup
                        %
                        %
                        \pgfmathfloatcreate{1}{1.0}{0}
                        \let\ONE=\pgfmathresult
                        \ifx\result\ONE
                            \pgfkeysalso{/pgfplots/small value}
                        \else
                            \pgfkeysalso{/pgfplots/large value}
                        \fi
                    }
                },
                check for zero,
            },
        },
        nodes near coords black white=0.5,
    ]
        \addplot[
            matrix plot,
            mesh/cols=2,
            point meta=explicit,draw=gray
        ] table [meta=C] {
            x y C
0 0 0.963
1 0 0.037
0 1 0.042
1 1 0.958
         };
    \end{axis}
\end{tikzpicture}
\end{adjustbox}
\\[-0.5ex]
\adjustbox{scale=0.9}{(d) MLP}
&
\adjustbox{scale=0.9}{(e) DNN}
&
\adjustbox{scale=0.9}{(f) LSTM}
\end{tabular}
\caption{Baseline confusion matrices (GloVe embeddings)}\label{fig:conf_baseline}
\end{figure}
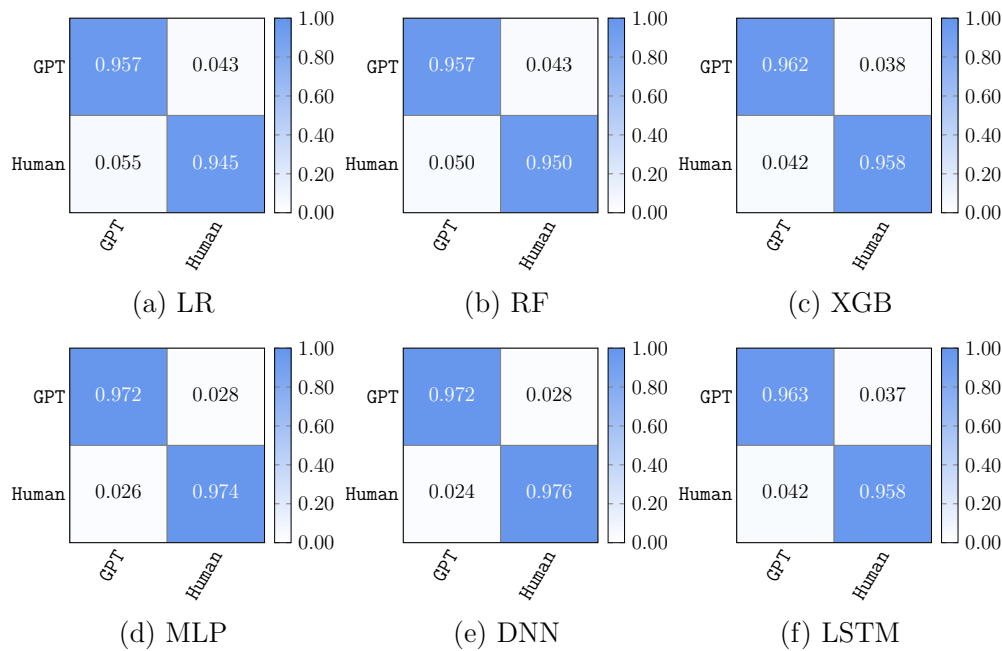

Figure~\ref{fig:conf_baseline} gives confusion matrices for all classification models in the 
baseline case, based on GloVe embeddings.
We observe that the misclassifications of human text as GPT text and the 
misclassifications of GPT-generated text as human
are nearly balanced in all cases. As with all three embedding
techniques, for this baseline case,
DNN performs best, with MLP being a close second.

\subsection{Baseline Models Tested on Transformed Text}

In this case, we first generate a transformed GPT-generated dataset by passing the data 
through our fine-tuned T5-small Seq2Seq model, and we generate a second transformed dataset by
passing the GPT-generated data through our fine-tuned BART Seq2Seq model.
We then test each of the six classification models trained in the baseline case
on these transformed samples, with the human-generated data again serving as the other class
in each of these binary classification experiment. Table~\ref{tab:transform-all} 
summarizes the results obtained for all of these classification experiments.

\begin{table}[!htb]
\centering\renewcommand{\arraystretch}{0.95}
\caption{Accuracy on transformed text}\label{tab:transform-all}
\adjustbox{scale=0.8}{
\begin{tabular}{c|ccc}\toprule
\multirow{2}{*}{\textbf{Embedding}} &
\multirow{2}{*}{\textbf{Model}} & \multicolumn{2}{c}{\textbf{Accuracy}} \\ 
 & & \textbf{T5-small} & \textbf{BART} \\ \midrule
\multirow{6}{*}{Word2Vec} 
& LR & 0.7650 & 0.7710 \\ 
& RF & 0.7505 & 0.7680 \\ 
& XGB & 0.7830 & 0.7590 \\ 
& MLP & 0.7860 & 0.7730 \\ 
& DNN & \textbf{0.7890} & \textbf{0.7950} \\ 
& LSTM & 0.7515 & 0.7630 \\  \midrule
\multirow{6}{*}{GloVe} 
& LR & 0.7840 & 0.7880 \\ 
& RF & 0.7910 & 0.7950 \\ 
& XGB & 0.8060 & 0.7840 \\ 
& MLP & 0.8275 & 0.8150 \\ 
& DNN & \textbf{0.8280} & \textbf{0.8220} \\ 
& LSTM & 0.7990 & 0.8010 \\  \midrule
\multirow{6}{*}{BERT} 
& LR & 0.8385 & 0.8715 \\ 
& RF & 0.8330 & 0.8685 \\ 
& XGB & 0.8420 & 0.8720 \\ 
& MLP & 0.8520 & 0.8860 \\ 
& DNN & \textbf{0.8940} & \textbf{0.9080} \\ 
& LSTM & 0.7195 & 0.7295 \\  \bottomrule
\end{tabular}
}
\end{table}

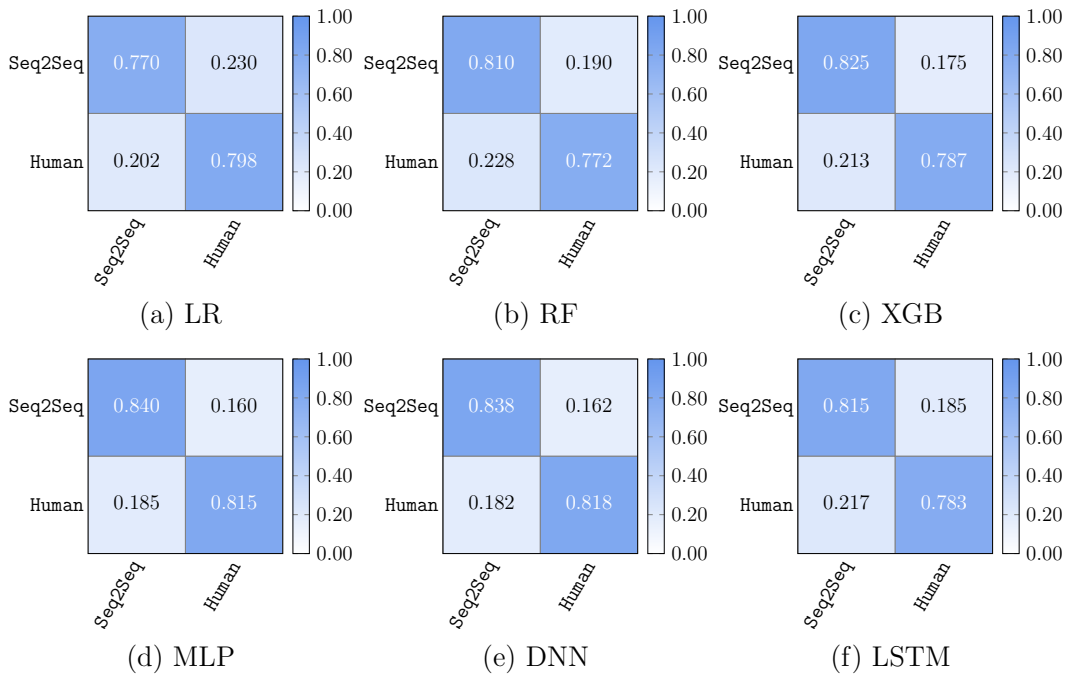
\begin{figure}[!htb]
\advance\tabcolsep by-6.5pt
\centering
\begin{tabular}{ccc}
\begin{adjustbox}{scale=0.875}
\begin{tikzpicture}[scale=0.5]
    \begin{axis}[
        width=7.5cm,
        height=7.5cm,
	colormap={bluewhite}{color=(white) rgb255=(100,149,237)},
        xticklabels={
        \texttt{Seq2Seq},
        \texttt{Human},
        },
        xtick={0,...,1},
        xtick style={draw=none},
	xticklabel style={anchor=east,rotate=60,yshift=-5pt,font=\tt,scale=1.5},
        yticklabels={
        \texttt{Seq2Seq},
        \texttt{Human},
        },
        ytick={0,...,1},
        ytick style={draw=none},
        enlargelimits=false,
        yticklabel style={xshift=2.5pt,font=\tt,scale=1.5},
        colorbar,
        colorbar style={
            ytick={0.00,0.20,0.40,0.60,0.80,1.00},
            yticklabels={0.00,0.20,0.40,0.60,0.80,1.00},
            yticklabel={\pgfmathprintnumber\tick},
            yticklabel style={font=\tt, scale=1.5,
            		/pgf/number format/fixed,
			/pgf/number format/fixed zerofill,
			/pgf/number format/precision=2}
        },
        point meta min=0.0,
        point meta max=1.0,
        nodes near coords={\pgfmathprintnumber\pgfplotspointmeta},
        nodes near coords black white/.style={
            small value/.style={
                yshift=-10pt,
                text=black,
                /pgf/number format/fixed,
                /pgf/number format/precision=3,
                /pgf/number format/zerofill=true,
                scale=1.5,
            },
            large value/.style={
                yshift=-10pt,
                text=white,
                /pgf/number format/fixed,
                /pgf/number format/precision=3,
                /pgf/number format/zerofill=true,
                scale=1.5,
            },
            every node near coord/.style={
                check for zero/.code={
                    \pgfmathfloatifflags{\pgfplotspointmeta}{0}{
                        \pgfkeys{/tikz/coordinate}
                    }{
                        \begingroup
                        \pgfkeys{/pgf/fpu}
                        \pgfmathparse{\pgfplotspointmeta<#1}
                        \global\let\result=\pgfmathresult
                        \endgroup
                        %
                        %
                        \pgfmathfloatcreate{1}{1.0}{0}
                        \let\ONE=\pgfmathresult
                        \ifx\result\ONE
                            \pgfkeysalso{/pgfplots/small value}
                        \else
                            \pgfkeysalso{/pgfplots/large value}
                        \fi
                    }
                },
                check for zero,
            },
        },
        nodes near coords black white=0.5,
    ]
        \addplot[
            matrix plot,
            mesh/cols=2,
            point meta=explicit,draw=gray
        ] table [meta=C] {
            x y C
0 0 0.770
1 0 0.230
0 1 0.202
1 1 0.798
         };
    \end{axis}
\end{tikzpicture}
\end{adjustbox}
&
\begin{adjustbox}{scale=0.875}
\begin{tikzpicture}[scale=0.5]
    \begin{axis}[
        width=7.5cm,
        height=7.5cm,
	colormap={bluewhite}{color=(white) rgb255=(100,149,237)},
        xticklabels={
        \texttt{Seq2Seq},
        \texttt{Human},
        },
        xtick={0,...,1},
        xtick style={draw=none},
	xticklabel style={anchor=east,rotate=60,yshift=-5pt,font=\tt,scale=1.5},
        yticklabels={
        \texttt{Seq2Seq},
        \texttt{Human},
        },
        ytick={0,...,1},
        ytick style={draw=none},
        enlargelimits=false,
        yticklabel style={xshift=2.5pt,font=\tt,scale=1.5},
        colorbar,
        colorbar style={
            ytick={0.00,0.20,0.40,0.60,0.80,1.00},
            yticklabels={0.00,0.20,0.40,0.60,0.80,1.00},
            yticklabel={\pgfmathprintnumber\tick},
            yticklabel style={font=\tt, scale=1.5,
            		/pgf/number format/fixed,
			/pgf/number format/fixed zerofill,
			/pgf/number format/precision=2}
        },
        point meta min=0.0,
        point meta max=1.0,
        nodes near coords={\pgfmathprintnumber\pgfplotspointmeta},
        nodes near coords black white/.style={
            small value/.style={
                yshift=-10pt,
                text=black,
                /pgf/number format/fixed,
                /pgf/number format/precision=3,
                /pgf/number format/zerofill=true,
                scale=1.5,
            },
            large value/.style={
                yshift=-10pt,
                text=white,
                /pgf/number format/fixed,
                /pgf/number format/precision=3,
                /pgf/number format/zerofill=true,
                scale=1.5,
            },
            every node near coord/.style={
                check for zero/.code={
                    \pgfmathfloatifflags{\pgfplotspointmeta}{0}{
                        \pgfkeys{/tikz/coordinate}
                    }{
                        \begingroup
                        \pgfkeys{/pgf/fpu}
                        \pgfmathparse{\pgfplotspointmeta<#1}
                        \global\let\result=\pgfmathresult
                        \endgroup
                        %
                        %
                        \pgfmathfloatcreate{1}{1.0}{0}
                        \let\ONE=\pgfmathresult
                        \ifx\result\ONE
                            \pgfkeysalso{/pgfplots/small value}
                        \else
                            \pgfkeysalso{/pgfplots/large value}
                        \fi
                    }
                },
                check for zero,
            },
        },
        nodes near coords black white=0.5,
    ]
        \addplot[
            matrix plot,
            mesh/cols=2,
            point meta=explicit,draw=gray
        ] table [meta=C] {
            x y C
0 0 0.810
1 0 0.190
0 1 0.228
1 1 0.772
         };
    \end{axis}
\end{tikzpicture}
\end{adjustbox}
&
\begin{adjustbox}{scale=0.875}
\begin{tikzpicture}[scale=0.5]
    \begin{axis}[
        width=7.5cm,
        height=7.5cm,
	colormap={bluewhite}{color=(white) rgb255=(100,149,237)},
        xticklabels={
        \texttt{Seq2Seq},
        \texttt{Human},
        },
        xtick={0,...,1},
        xtick style={draw=none},
	xticklabel style={anchor=east,rotate=60,yshift=-5pt,font=\tt,scale=1.5},
        yticklabels={
        \texttt{Seq2Seq},
        \texttt{Human},
        },
        ytick={0,...,1},
        ytick style={draw=none},
        enlargelimits=false,
        yticklabel style={xshift=2.5pt,font=\tt,scale=1.5},
        colorbar,
        colorbar style={
            ytick={0.00,0.20,0.40,0.60,0.80,1.00},
            yticklabels={0.00,0.20,0.40,0.60,0.80,1.00},
            yticklabel={\pgfmathprintnumber\tick},
            yticklabel style={font=\tt, scale=1.5,
            		/pgf/number format/fixed,
			/pgf/number format/fixed zerofill,
			/pgf/number format/precision=2}
        },
        point meta min=0.0,
        point meta max=1.0,
        nodes near coords={\pgfmathprintnumber\pgfplotspointmeta},
        nodes near coords black white/.style={
            small value/.style={
                yshift=-10pt,
                text=black,
                /pgf/number format/fixed,
                /pgf/number format/precision=3,
                /pgf/number format/zerofill=true,
                scale=1.5,
            },
            large value/.style={
                yshift=-10pt,
                text=white,
                /pgf/number format/fixed,
                /pgf/number format/precision=3,
                /pgf/number format/zerofill=true,
                scale=1.5,
            },
            every node near coord/.style={
                check for zero/.code={
                    \pgfmathfloatifflags{\pgfplotspointmeta}{0}{
                        \pgfkeys{/tikz/coordinate}
                    }{
                        \begingroup
                        \pgfkeys{/pgf/fpu}
                        \pgfmathparse{\pgfplotspointmeta<#1}
                        \global\let\result=\pgfmathresult
                        \endgroup
                        %
                        %
                        \pgfmathfloatcreate{1}{1.0}{0}
                        \let\ONE=\pgfmathresult
                        \ifx\result\ONE
                            \pgfkeysalso{/pgfplots/small value}
                        \else
                            \pgfkeysalso{/pgfplots/large value}
                        \fi
                    }
                },
                check for zero,
            },
        },
        nodes near coords black white=0.5,
    ]
        \addplot[
            matrix plot,
            mesh/cols=2,
            point meta=explicit,draw=gray
        ] table [meta=C] {
            x y C
0 0 0.825
1 0 0.175
0 1 0.213
1 1 0.787
         };
    \end{axis}
\end{tikzpicture}
\end{adjustbox}
\\[-1ex]
\adjustbox{scale=0.9}{(a) LR}
&
\adjustbox{scale=0.9}{(b) RF}
&
\adjustbox{scale=0.9}{(c) XGB}
\\ \\[-2ex]
\begin{adjustbox}{scale=0.875}
\begin{tikzpicture}[scale=0.5]
    \begin{axis}[
        width=7.5cm,
        height=7.5cm,
	colormap={bluewhite}{color=(white) rgb255=(100,149,237)},
        xticklabels={
        \texttt{Seq2Seq},
        \texttt{Human},
        },
        xtick={0,...,1},
        xtick style={draw=none},
	xticklabel style={anchor=east,rotate=60,yshift=-5pt,font=\tt,scale=1.5},
        yticklabels={
        \texttt{Seq2Seq},
        \texttt{Human},
        },
        ytick={0,...,1},
        ytick style={draw=none},
        enlargelimits=false,
        yticklabel style={xshift=2.5pt,font=\tt,scale=1.5},
        colorbar,
        colorbar style={
            ytick={0.00,0.20,0.40,0.60,0.80,1.00},
            yticklabels={0.00,0.20,0.40,0.60,0.80,1.00},
            yticklabel={\pgfmathprintnumber\tick},
            yticklabel style={font=\tt, scale=1.5,
            		/pgf/number format/fixed,
			/pgf/number format/fixed zerofill,
			/pgf/number format/precision=2}
        },
        point meta min=0.0,
        point meta max=1.0,
        nodes near coords={\pgfmathprintnumber\pgfplotspointmeta},
        nodes near coords black white/.style={
            small value/.style={
                yshift=-10pt,
                text=black,
                /pgf/number format/fixed,
                /pgf/number format/precision=3,
                /pgf/number format/zerofill=true,
                scale=1.5,
            },
            large value/.style={
                yshift=-10pt,
                text=white,
                /pgf/number format/fixed,
                /pgf/number format/precision=3,
                /pgf/number format/zerofill=true,
                scale=1.5,
            },
            every node near coord/.style={
                check for zero/.code={
                    \pgfmathfloatifflags{\pgfplotspointmeta}{0}{
                        \pgfkeys{/tikz/coordinate}
                    }{
                        \begingroup
                        \pgfkeys{/pgf/fpu}
                        \pgfmathparse{\pgfplotspointmeta<#1}
                        \global\let\result=\pgfmathresult
                        \endgroup
                        %
                        %
                        \pgfmathfloatcreate{1}{1.0}{0}
                        \let\ONE=\pgfmathresult
                        \ifx\result\ONE
                            \pgfkeysalso{/pgfplots/small value}
                        \else
                            \pgfkeysalso{/pgfplots/large value}
                        \fi
                    }
                },
                check for zero,
            },
        },
        nodes near coords black white=0.5,
    ]
        \addplot[
            matrix plot,
            mesh/cols=2,
            point meta=explicit,draw=gray
        ] table [meta=C] {
            x y C
0 0 0.840
1 0 0.160
0 1 0.185
1 1 0.815
         };
    \end{axis}
\end{tikzpicture}
\end{adjustbox}
&
\begin{adjustbox}{scale=0.875}
\begin{tikzpicture}[scale=0.5]
    \begin{axis}[
        width=7.5cm,
        height=7.5cm,
	colormap={bluewhite}{color=(white) rgb255=(100,149,237)},
        xticklabels={
        \texttt{Seq2Seq},
        \texttt{Human},
        },
        xtick={0,...,1},
        xtick style={draw=none},
	xticklabel style={anchor=east,rotate=60,yshift=-5pt,font=\tt,scale=1.5},
        yticklabels={
        \texttt{Seq2Seq},
        \texttt{Human},
        },
        ytick={0,...,1},
        ytick style={draw=none},
        enlargelimits=false,
        yticklabel style={xshift=2.5pt,font=\tt,scale=1.5},
        colorbar,
        colorbar style={
            ytick={0.00,0.20,0.40,0.60,0.80,1.00},
            yticklabels={0.00,0.20,0.40,0.60,0.80,1.00},
            yticklabel={\pgfmathprintnumber\tick},
            yticklabel style={font=\tt, scale=1.5,
            		/pgf/number format/fixed,
			/pgf/number format/fixed zerofill,
			/pgf/number format/precision=2}
        },
        point meta min=0.0,
        point meta max=1.0,
        nodes near coords={\pgfmathprintnumber\pgfplotspointmeta},
        nodes near coords black white/.style={
            small value/.style={
                yshift=-10pt,
                text=black,
                /pgf/number format/fixed,
                /pgf/number format/precision=3,
                /pgf/number format/zerofill=true,
                scale=1.5,
            },
            large value/.style={
                yshift=-10pt,
                text=white,
                /pgf/number format/fixed,
                /pgf/number format/precision=3,
                /pgf/number format/zerofill=true,
                scale=1.5,
            },
            every node near coord/.style={
                check for zero/.code={
                    \pgfmathfloatifflags{\pgfplotspointmeta}{0}{
                        \pgfkeys{/tikz/coordinate}
                    }{
                        \begingroup
                        \pgfkeys{/pgf/fpu}
                        \pgfmathparse{\pgfplotspointmeta<#1}
                        \global\let\result=\pgfmathresult
                        \endgroup
                        %
                        %
                        \pgfmathfloatcreate{1}{1.0}{0}
                        \let\ONE=\pgfmathresult
                        \ifx\result\ONE
                            \pgfkeysalso{/pgfplots/small value}
                        \else
                            \pgfkeysalso{/pgfplots/large value}
                        \fi
                    }
                },
                check for zero,
            },
        },
        nodes near coords black white=0.5,
    ]
        \addplot[
            matrix plot,
            mesh/cols=2,
            point meta=explicit,draw=gray
        ] table [meta=C] {
            x y C
0 0 0.838
1 0 0.162
0 1 0.182
1 1 0.818
         };
    \end{axis}
\end{tikzpicture}
\end{adjustbox}
&
\begin{adjustbox}{scale=0.875}
\begin{tikzpicture}[scale=0.5]
    \begin{axis}[
        width=7.5cm,
        height=7.5cm,
	colormap={bluewhite}{color=(white) rgb255=(100,149,237)},
        xticklabels={
        \texttt{Seq2Seq},
        \texttt{Human},
        },
        xtick={0,...,1},
        xtick style={draw=none},
	xticklabel style={anchor=east,rotate=60,yshift=-5pt,font=\tt,scale=1.5},
        yticklabels={
        \texttt{Seq2Seq},
        \texttt{Human},
        },
        ytick={0,...,1},
        ytick style={draw=none},
        enlargelimits=false,
        yticklabel style={xshift=2.5pt,font=\tt,scale=1.5},
        colorbar,
        colorbar style={
            ytick={0.00,0.20,0.40,0.60,0.80,1.00},
            yticklabels={0.00,0.20,0.40,0.60,0.80,1.00},
            yticklabel={\pgfmathprintnumber\tick},
            yticklabel style={font=\tt, scale=1.5,
            		/pgf/number format/fixed,
			/pgf/number format/fixed zerofill,
			/pgf/number format/precision=2}
        },
        point meta min=0.0,
        point meta max=1.0,
        nodes near coords={\pgfmathprintnumber\pgfplotspointmeta},
        nodes near coords black white/.style={
            small value/.style={
                yshift=-10pt,
                text=black,
                /pgf/number format/fixed,
                /pgf/number format/precision=3,
                /pgf/number format/zerofill=true,
                scale=1.5,
            },
            large value/.style={
                yshift=-10pt,
                text=white,
                /pgf/number format/fixed,
                /pgf/number format/precision=3,
                /pgf/number format/zerofill=true,
                scale=1.5,
            },
            every node near coord/.style={
                check for zero/.code={
                    \pgfmathfloatifflags{\pgfplotspointmeta}{0}{
                        \pgfkeys{/tikz/coordinate}
                    }{
                        \begingroup
                        \pgfkeys{/pgf/fpu}
                        \pgfmathparse{\pgfplotspointmeta<#1}
                        \global\let\result=\pgfmathresult
                        \endgroup
                        %
                        %
                        \pgfmathfloatcreate{1}{1.0}{0}
                        \let\ONE=\pgfmathresult
                        \ifx\result\ONE
                            \pgfkeysalso{/pgfplots/small value}
                        \else
                            \pgfkeysalso{/pgfplots/large value}
                        \fi
                    }
                },
                check for zero,
            },
        },
        nodes near coords black white=0.5,
    ]
        \addplot[
            matrix plot,
            mesh/cols=2,
            point meta=explicit,draw=gray
        ] table [meta=C] {
            x y C
0 0 0.815
1 0 0.185
0 1 0.217
1 1 0.783
         };
    \end{axis}
\end{tikzpicture}
\end{adjustbox}
\\[-1ex]
\adjustbox{scale=0.9}{(d) MLP}
&
\adjustbox{scale=0.9}{(e) DNN}
&
\adjustbox{scale=0.9}{(f) LSTM}
\end{tabular}
\caption{Confusion matrices after transformation (GloVe and T5-small)}\label{fig:conf_after}
\end{figure}

Figure~\ref{fig:conf_after} gives confusion matrices for all classification models in the 
case where T5-small is used to transform the GPT-generate text, based on GloVe embeddings.
All models except LR have more misclassifications of human text as GPT text than the converse.

\begin{figure}[!htb]
    \centering\advance\tabcolsep by -2.5pt
    \begin{tabular}{cc}
    \begin{tikzpicture}[scale=0.9, every node/.style={scale=1.0}]
\pgfkeys{/pgf/number format/.cd,1000 sep={}}
\begin{axis}[
        width  = 0.52*\textwidth,
        height = 5.75cm,
        ymin=0.0,ymax=1.3,
        ytick={0.0, 0.2, 0.4, 0.6, 0.8, 1.0},
        major x tick style = transparent,
        ybar=5*\pgflinewidth,
        bar width=6.5pt,
        ylabel = {Accuracy},
        ylabel style = {scale=0.85},
        symbolic x coords={
        		LR, RF, XGB, MLP, DNN, LSTM
        },
        xticklabels={
        		LR, RF, XGB, MLP, DNN, LSTM
        },
	y tick label style={scale=0.80,
    		/pgf/number format/.cd,
   		fixed,
   		fixed zerofill,
    		precision=2},
        xtick = data,
        x tick label style={scale=0.8,
		},
        nodes near coords,
        every node near coord/.append style={rotate=90, scale=0.7,
        								   anchor=west, 
								   /pgf/number format/.cd,
								   fixed,
								   fixed zerofill,
								   precision=4},
        enlarge x limits=0.1125,
        legend cell align=left,
        legend pos=south east,
        legend style={nodes={scale=0.85},
        },
]
\addplot [fill=blue,opacity=1.00]
coordinates {
(LR, 0.9300)
(RF, 0.9205)
(XGB, 0.9345)
(MLP, 0.9395)
(DNN, 0.9490)
(LSTM, 0.9315)
};
\addlegendentry{Baseline}
\addplot [fill=red,opacity=1.00]
coordinates {
(LR, 0.7650)
(RF, 0.7505)
(XGB, 0.7830)
(MLP, 0.7860)
(DNN, 0.7890)
(LSTM, 0.7515)
};
\addlegendentry{T5-small}
\addplot [fill=yellow,opacity=1.00]
coordinates {
(LR, 0.7710)
(RF, 0.7680)
(XGB, 0.7590)
(MLP, 0.7730)
(DNN, 0.7950)
(LSTM, 0.7630)
};
\addlegendentry{BART}
\end{axis}
\end{tikzpicture}
    &
    \begin{tikzpicture}[scale=0.9, every node/.style={scale=1.0}]
\pgfkeys{/pgf/number format/.cd,1000 sep={}}
\begin{axis}[
        width  = 0.52*\textwidth,
        height = 5.75cm,
        ymin=0.0,ymax=1.3,
        ytick={0.0, 0.2, 0.4, 0.6, 0.8, 1.0},
        major x tick style = transparent,
        ybar=5*\pgflinewidth,
        bar width=6.5pt,
        symbolic x coords={
        		LR, RF, XGB, MLP, DNN, LSTM
        },
        xticklabels={
        		LR, RF, XGB, MLP, DNN, LSTM
        },
	y tick label style={scale=0.80,
    		/pgf/number format/.cd,
   		fixed,
   		fixed zerofill,
    		precision=2},
        xtick = data,
        x tick label style={scale=0.8,
		},
        nodes near coords,
        every node near coord/.append style={rotate=90, scale=0.7,
        								   anchor=west, 
								   /pgf/number format/.cd,
								   fixed,
								   fixed zerofill,
								   precision=4},
        enlarge x limits=0.1125,
        legend cell align=left,
        legend pos=south east,
        legend style={nodes={scale=0.85},
        },
]
\addplot [fill=blue,opacity=1.00]
coordinates {
(LR, 0.9510)
(RF, 0.9535)
(XGB, 0.9600)
(MLP, 0.9730)
(DNN, 0.9740)
(LSTM, 0.9605)
};
\addlegendentry{Baseline}
\addplot [fill=red,opacity=1.00]
coordinates {
(LR, 0.7840)
(RF, 0.7910)
(XGB, 0.8060)
(MLP, 0.8275)
(DNN, 0.8280)
(LSTM, 0.7990)
};
\addlegendentry{T5-small}
\addplot [fill=yellow,opacity=1.00]
coordinates {
(LR, 0.7880)
(RF, 0.7950)
(XGB, 0.7840)
(MLP, 0.8150)
(DNN, 0.8220)
(LSTM, 0.8010)
};
\addlegendentry{BART}
\end{axis}
\end{tikzpicture}
    \\
    \adjustbox{scale=0.9}{(a) Word2Vec}
    &
    \adjustbox{scale=0.9}{(b) GloVe}
    \\ \\[-1.25ex]
    \multicolumn{2}{c}{\begin{tikzpicture}[scale=0.9, every node/.style={scale=1.0}]
\pgfkeys{/pgf/number format/.cd,1000 sep={}}
\begin{axis}[
        width  = 0.52*\textwidth,
        height = 5.75cm,
        ymin=0.0,ymax=1.3,
        ytick={0.0, 0.2, 0.4, 0.6, 0.8, 1.0},
        major x tick style = transparent,
        ybar=5*\pgflinewidth,
        bar width=6.5pt,
        ylabel = {Accuracy},
        ylabel style = {scale=0.85},
        symbolic x coords={
        		LR, RF, XGB, MLP, DNN, LSTM
        },
        xticklabels={
        		LR, RF, XGB, MLP, DNN, LSTM
        },
	y tick label style={scale=0.80,
    		/pgf/number format/.cd,
   		fixed,
   		fixed zerofill,
    		precision=2},
        xtick = data,
        x tick label style={scale=0.8,
		},
        nodes near coords,
        every node near coord/.append style={rotate=90, scale=0.7,
        								   anchor=west, 
								   /pgf/number format/.cd,
								   fixed,
								   fixed zerofill,
								   precision=4},
        enlarge x limits=0.1125,
        legend cell align=left,
        legend pos=south east,
        legend style={nodes={scale=0.85},
        },
]
\addplot [fill=blue,opacity=1.00]
coordinates {
(LR, 0.9825)
(RF, 0.9585)
(XGB, 0.9720)
(MLP, 0.9820)
(DNN, 0.9840)
(LSTM, 0.8495)
};
\addlegendentry{Baseline}
\addplot [fill=red,opacity=1.00]
coordinates {
(LR, 0.8385)
(RF, 0.8330)
(XGB, 0.8420)
(MLP, 0.8520)
(DNN, 0.8940)
(LSTM, 0.7195)
};
\addlegendentry{T5-small}
\addplot [fill=yellow,opacity=1.00]
coordinates {
(LR, 0.8715)
(RF, 0.8685)
(XGB, 0.8720)
(MLP, 0.8860)
(DNN, 0.9080)
(LSTM, 0.7295)
};
\addlegendentry{BART}
\end{axis}
\end{tikzpicture}}
    \\
    \multicolumn{2}{c}{\adjustbox{scale=0.9}{(c) BERT}}
    \end{tabular}
    \caption{Accuracy comparison for transformed text}
    \label{fig:accuracy-comp}
\end{figure}
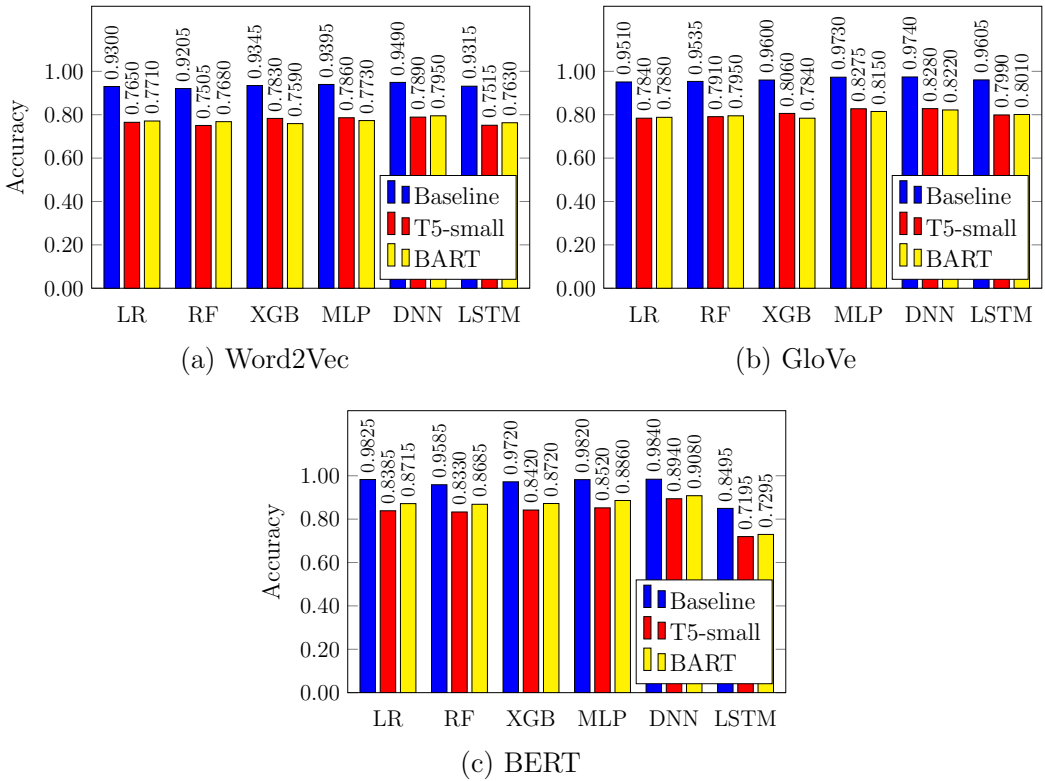

For ease of comparison, we present the results from 
Tables~\ref{tab:baseline-all} and~\ref{tab:transform-all}
in bar graph form in Figure~\ref{fig:accuracy-comp}.
We observe that for Word2Vec embedding 
there is a consistent decrease in accuracy, with the
minimum decrease being for the XGB model and T5-small transformation
which yields a change of
$$
     \frac{0.9345-0.7830}{0.9345} = 16.2\%
$$
On the other hand, the maximum decrease in accuracy occurs for
the LSTM model and T5-small transformation, which yields 
a change of
$$
    \frac{0.9315-0.7515}{0.9315} = 19.3\%
$$
For four of the six classifiers, T5-transformed data 
results in a slightly greater drop in accuracy, as compared to BART. 

For GloVe embeddings, a similar trend is observed, with the accuracy 
decrease being in the range of~14.9\%\ (MLP and T5-small) to~18.3\%\ (XGB and BART). 
In this case,
three of the models have a bigger decrease in accuracy for T5-transformed
data while the remaining three show a larger drop for BART-transformed data.

BERT embeddings are more difficult to defeat, as the
decrease in accuracy ranges from~7.7\%\ (DNN and BART) 
to~15.3\%\ (LSTM and T5-small).
For BERT, the T5-transformed data yields a larger drop in 
accuracy for every model.

Overall, the DNN model consistently performs best
for each of the three embedding techniques, and the
results for BERT embeddings are measurably better than
for Word2Vec or GloVe. The most robust case consists of
BERT embeddings and the BART-transformed data,
which yields an accuracy of~0.9080. Assuming that 
the DNN model is used, an attacker would
choose to transform the GPT-generated data using the T5-small Seq2Seq 
model, which would result in a decline of~9.1\%\ in classification accuracy.
For all other classification models tested, an attacker would be able to 
produce a larger drop in accuracy. 



\subsection{Models Retrained on Transformed Text}

As a final set of experiments, we retrain and test our classification models 
on Seq2Seq-transformed data versus human-generated text.
As in Section~\ref{sect:case1}, we train and test each of six classification models
with each of the three word embedding techniques.
Note that we train each of these combinations on the \hbox{T5-transformed}
data, and we again train each combination of classification model and
embedding technique on the BART-transformed data.
The results of these experiments are summarized in Table~\ref{tab:retrain-all}, 
where we observe that the accuracy has increased for all models,
as compared to the results in the previous section.

\begin{table}[!htb]
\centering\renewcommand{\arraystretch}{0.95}
\caption{Accuracy after retraining\label{tab:retrain-all}}
\adjustbox{scale=0.8}{
\begin{tabular}{c|ccc}\toprule
\multirow{2}{*}{\textbf{Embedding}} & \multirow{2}{*}{\textbf{Model}} & 
\multicolumn{2}{c}{\textbf{Accuracy}} \\ 
 & & \textbf{T5-small} & \textbf{BART} \\ \midrule
\multirow{6}{*}{Word2Vec}
& LR & 0.9355 & 0.9280 \\ 
& RF & 0.9235 & 0.9210 \\ 
& XGB & 0.9370 & 0.9360 \\ 
& MLP & 0.9410 & 0.9365 \\ 
& DNN & \textbf{0.9520} & \textbf{0.9500} \\ 
& LSTM & 0.9340 & 0.9290 \\  \midrule
\multirow{6}{*}{GloVe}
& LR & 0.9530 & 0.9465 \\ 
& RF & 0.9550 & 0.9530 \\ 
& XGB & 0.9625 & 0.9540 \\ 
& MLP & 0.9640 & 0.9610 \\ 
& DNN & \textbf{0.9760} & \textbf{0.9680} \\ 
& LSTM & 0.9615 & 0.9570 \\  \midrule
\multirow{6}{*}{BERT}
& LR & 0.9740 & 0.9710 \\ 
& RF & 0.9605 & 0.9585 \\ 
& XGB & 0.9735 & 0.9730 \\ 
& MLP & 0.9805 & 0.9810 \\ 
& DNN & \textbf{0.9845} & \textbf{0.9825} \\
& LSTM & 0.8510 & 0.8390 \\  \bottomrule
\end{tabular}
}
\end{table}

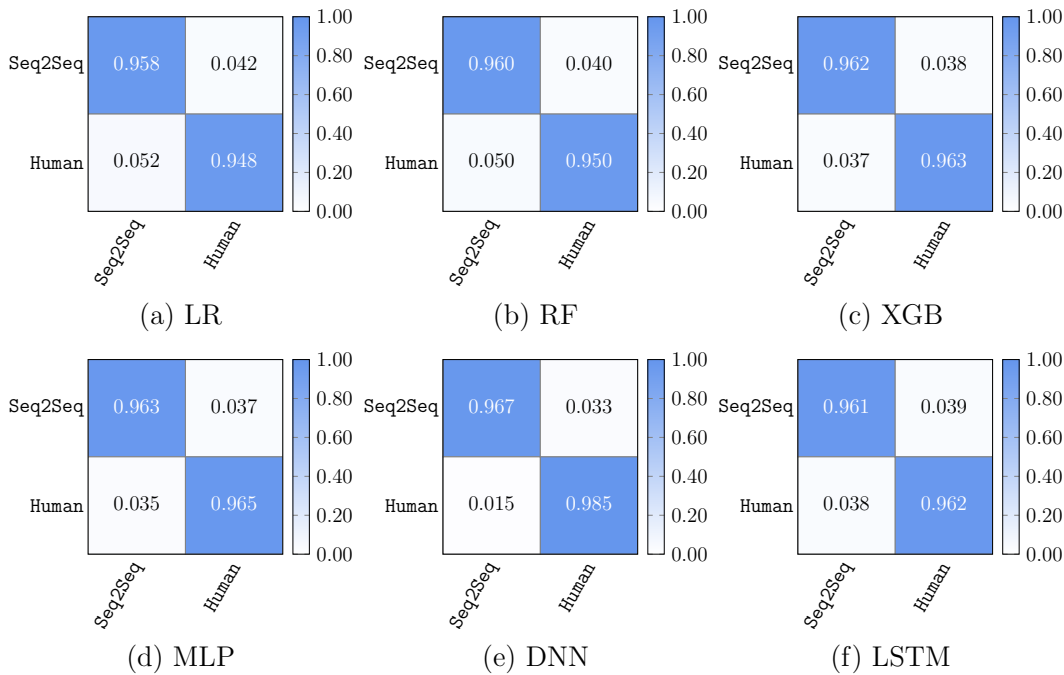
\begin{figure}[!htb]
\advance\tabcolsep by-6.5pt
\centering
\begin{tabular}{ccc}
\begin{adjustbox}{scale=0.875}
\begin{tikzpicture}[scale=0.5]
    \begin{axis}[
        width=7.5cm,
        height=7.5cm,
	colormap={bluewhite}{color=(white) rgb255=(100,149,237)},
        xticklabels={
        \texttt{Seq2Seq},
        \texttt{Human},
        },
        xtick={0,...,1},
        xtick style={draw=none},
	xticklabel style={anchor=east,rotate=60,yshift=-5pt,font=\tt,scale=1.5},
        yticklabels={
        \texttt{Seq2Seq},
        \texttt{Human},
        },
        ytick={0,...,1},
        ytick style={draw=none},
        enlargelimits=false,
        yticklabel style={xshift=2.5pt,font=\tt,scale=1.5},
        colorbar,
        colorbar style={
            ytick={0.00,0.20,0.40,0.60,0.80,1.00},
            yticklabels={0.00,0.20,0.40,0.60,0.80,1.00},
            yticklabel={\pgfmathprintnumber\tick},
            yticklabel style={font=\tt, scale=1.5,
            		/pgf/number format/fixed,
			/pgf/number format/fixed zerofill,
			/pgf/number format/precision=2}
        },
        point meta min=0.0,
        point meta max=1.0,
        nodes near coords={\pgfmathprintnumber\pgfplotspointmeta},
        nodes near coords black white/.style={
            small value/.style={
                yshift=-10pt,
                text=black,
                /pgf/number format/fixed,
                /pgf/number format/precision=3,
                /pgf/number format/zerofill=true,
                scale=1.5,
            },
            large value/.style={
                yshift=-10pt,
                text=white,
                /pgf/number format/fixed,
                /pgf/number format/precision=3,
                /pgf/number format/zerofill=true,
                scale=1.5,
            },
            every node near coord/.style={
                check for zero/.code={
                    \pgfmathfloatifflags{\pgfplotspointmeta}{0}{
                        \pgfkeys{/tikz/coordinate}
                    }{
                        \begingroup
                        \pgfkeys{/pgf/fpu}
                        \pgfmathparse{\pgfplotspointmeta<#1}
                        \global\let\result=\pgfmathresult
                        \endgroup
                        %
                        %
                        \pgfmathfloatcreate{1}{1.0}{0}
                        \let\ONE=\pgfmathresult
                        \ifx\result\ONE
                            \pgfkeysalso{/pgfplots/small value}
                        \else
                            \pgfkeysalso{/pgfplots/large value}
                        \fi
                    }
                },
                check for zero,
            },
        },
        nodes near coords black white=0.5,
    ]
        \addplot[
            matrix plot,
            mesh/cols=2,
            point meta=explicit,draw=gray
        ] table [meta=C] {
            x y C
0 0 0.958
1 0 0.042
0 1 0.052
1 1 0.948
         };
    \end{axis}
\end{tikzpicture}
\end{adjustbox}
&
\begin{adjustbox}{scale=0.875}
\begin{tikzpicture}[scale=0.5]
    \begin{axis}[
        width=7.5cm,
        height=7.5cm,
	colormap={bluewhite}{color=(white) rgb255=(100,149,237)},
        xticklabels={
        \texttt{Seq2Seq},
        \texttt{Human},
        },
        xtick={0,...,1},
        xtick style={draw=none},
	xticklabel style={anchor=east,rotate=60,yshift=-5pt,font=\tt,scale=1.5},
        yticklabels={
        \texttt{Seq2Seq},
        \texttt{Human},
        },
        ytick={0,...,1},
        ytick style={draw=none},
        enlargelimits=false,
        yticklabel style={xshift=2.5pt,font=\tt,scale=1.5},
        colorbar,
        colorbar style={
            ytick={0.00,0.20,0.40,0.60,0.80,1.00},
            yticklabels={0.00,0.20,0.40,0.60,0.80,1.00},
            yticklabel={\pgfmathprintnumber\tick},
            yticklabel style={font=\tt, scale=1.5,
            		/pgf/number format/fixed,
			/pgf/number format/fixed zerofill,
			/pgf/number format/precision=2}
        },
        point meta min=0.0,
        point meta max=1.0,
        nodes near coords={\pgfmathprintnumber\pgfplotspointmeta},
        nodes near coords black white/.style={
            small value/.style={
                yshift=-10pt,
                text=black,
                /pgf/number format/fixed,
                /pgf/number format/precision=3,
                /pgf/number format/zerofill=true,
                scale=1.5,
            },
            large value/.style={
                yshift=-10pt,
                text=white,
                /pgf/number format/fixed,
                /pgf/number format/precision=3,
                /pgf/number format/zerofill=true,
                scale=1.5,
            },
            every node near coord/.style={
                check for zero/.code={
                    \pgfmathfloatifflags{\pgfplotspointmeta}{0}{
                        \pgfkeys{/tikz/coordinate}
                    }{
                        \begingroup
                        \pgfkeys{/pgf/fpu}
                        \pgfmathparse{\pgfplotspointmeta<#1}
                        \global\let\result=\pgfmathresult
                        \endgroup
                        %
                        %
                        \pgfmathfloatcreate{1}{1.0}{0}
                        \let\ONE=\pgfmathresult
                        \ifx\result\ONE
                            \pgfkeysalso{/pgfplots/small value}
                        \else
                            \pgfkeysalso{/pgfplots/large value}
                        \fi
                    }
                },
                check for zero,
            },
        },
        nodes near coords black white=0.5,
    ]
        \addplot[
            matrix plot,
            mesh/cols=2,
            point meta=explicit,draw=gray
        ] table [meta=C] {
            x y C
0 0 0.960
1 0 0.040
0 1 0.050
1 1 0.950
         };
    \end{axis}
\end{tikzpicture}
\end{adjustbox}
&
\begin{adjustbox}{scale=0.875}
\begin{tikzpicture}[scale=0.5]
    \begin{axis}[
        width=7.5cm,
        height=7.5cm,
	colormap={bluewhite}{color=(white) rgb255=(100,149,237)},
        xticklabels={
        \texttt{Seq2Seq},
        \texttt{Human},
        },
        xtick={0,...,1},
        xtick style={draw=none},
	xticklabel style={anchor=east,rotate=60,yshift=-5pt,font=\tt,scale=1.5},
        yticklabels={
        \texttt{Seq2Seq},
        \texttt{Human},
        },
        ytick={0,...,1},
        ytick style={draw=none},
        enlargelimits=false,
        yticklabel style={xshift=2.5pt,font=\tt,scale=1.5},
        colorbar,
        colorbar style={
            ytick={0.00,0.20,0.40,0.60,0.80,1.00},
            yticklabels={0.00,0.20,0.40,0.60,0.80,1.00},
            yticklabel={\pgfmathprintnumber\tick},
            yticklabel style={font=\tt, scale=1.5,
            		/pgf/number format/fixed,
			/pgf/number format/fixed zerofill,
			/pgf/number format/precision=2}
        },
        point meta min=0.0,
        point meta max=1.0,
        nodes near coords={\pgfmathprintnumber\pgfplotspointmeta},
        nodes near coords black white/.style={
            small value/.style={
                yshift=-10pt,
                text=black,
                /pgf/number format/fixed,
                /pgf/number format/precision=3,
                /pgf/number format/zerofill=true,
                scale=1.5,
            },
            large value/.style={
                yshift=-10pt,
                text=white,
                /pgf/number format/fixed,
                /pgf/number format/precision=3,
                /pgf/number format/zerofill=true,
                scale=1.5,
            },
            every node near coord/.style={
                check for zero/.code={
                    \pgfmathfloatifflags{\pgfplotspointmeta}{0}{
                        \pgfkeys{/tikz/coordinate}
                    }{
                        \begingroup
                        \pgfkeys{/pgf/fpu}
                        \pgfmathparse{\pgfplotspointmeta<#1}
                        \global\let\result=\pgfmathresult
                        \endgroup
                        %
                        %
                        \pgfmathfloatcreate{1}{1.0}{0}
                        \let\ONE=\pgfmathresult
                        \ifx\result\ONE
                            \pgfkeysalso{/pgfplots/small value}
                        \else
                            \pgfkeysalso{/pgfplots/large value}
                        \fi
                    }
                },
                check for zero,
            },
        },
        nodes near coords black white=0.5,
    ]
        \addplot[
            matrix plot,
            mesh/cols=2,
            point meta=explicit,draw=gray
        ] table [meta=C] {
            x y C
0 0 0.962
1 0 0.038
0 1 0.037
1 1 0.963
         };
    \end{axis}
\end{tikzpicture}
\end{adjustbox}
\\[-1ex]
\adjustbox{scale=0.9}{(a) LR}
&
\adjustbox{scale=0.9}{(b) RF}
&
\adjustbox{scale=0.9}{(c) XGB}
\\ \\[-2ex]
\begin{adjustbox}{scale=0.875}
\begin{tikzpicture}[scale=0.5]
    \begin{axis}[
        width=7.5cm,
        height=7.5cm,
	colormap={bluewhite}{color=(white) rgb255=(100,149,237)},
        xticklabels={
        \texttt{Seq2Seq},
        \texttt{Human},
        },
        xtick={0,...,1},
        xtick style={draw=none},
	xticklabel style={anchor=east,rotate=60,yshift=-5pt,font=\tt,scale=1.5},
        yticklabels={
        \texttt{Seq2Seq},
        \texttt{Human},
        },
        ytick={0,...,1},
        ytick style={draw=none},
        enlargelimits=false,
        yticklabel style={xshift=2.5pt,font=\tt,scale=1.5},
        colorbar,
        colorbar style={
            ytick={0.00,0.20,0.40,0.60,0.80,1.00},
            yticklabels={0.00,0.20,0.40,0.60,0.80,1.00},
            yticklabel={\pgfmathprintnumber\tick},
            yticklabel style={font=\tt, scale=1.5,
            		/pgf/number format/fixed,
			/pgf/number format/fixed zerofill,
			/pgf/number format/precision=2}
        },
        point meta min=0.0,
        point meta max=1.0,
        nodes near coords={\pgfmathprintnumber\pgfplotspointmeta},
        nodes near coords black white/.style={
            small value/.style={
                yshift=-10pt,
                text=black,
                /pgf/number format/fixed,
                /pgf/number format/precision=3,
                /pgf/number format/zerofill=true,
                scale=1.5,
            },
            large value/.style={
                yshift=-10pt,
                text=white,
                /pgf/number format/fixed,
                /pgf/number format/precision=3,
                /pgf/number format/zerofill=true,
                scale=1.5,
            },
            every node near coord/.style={
                check for zero/.code={
                    \pgfmathfloatifflags{\pgfplotspointmeta}{0}{
                        \pgfkeys{/tikz/coordinate}
                    }{
                        \begingroup
                        \pgfkeys{/pgf/fpu}
                        \pgfmathparse{\pgfplotspointmeta<#1}
                        \global\let\result=\pgfmathresult
                        \endgroup
                        %
                        %
                        \pgfmathfloatcreate{1}{1.0}{0}
                        \let\ONE=\pgfmathresult
                        \ifx\result\ONE
                            \pgfkeysalso{/pgfplots/small value}
                        \else
                            \pgfkeysalso{/pgfplots/large value}
                        \fi
                    }
                },
                check for zero,
            },
        },
        nodes near coords black white=0.5,
    ]
        \addplot[
            matrix plot,
            mesh/cols=2,
            point meta=explicit,draw=gray
        ] table [meta=C] {
            x y C
0 0 0.963
1 0 0.037
0 1 0.035
1 1 0.965
         };
    \end{axis}
\end{tikzpicture}
\end{adjustbox}
&
\begin{adjustbox}{scale=0.875}
\begin{tikzpicture}[scale=0.5]
    \begin{axis}[
        width=7.5cm,
        height=7.5cm,
	colormap={bluewhite}{color=(white) rgb255=(100,149,237)},
        xticklabels={
        \texttt{Seq2Seq},
        \texttt{Human},
        },
        xtick={0,...,1},
        xtick style={draw=none},
	xticklabel style={anchor=east,rotate=60,yshift=-5pt,font=\tt,scale=1.5},
        yticklabels={
        \texttt{Seq2Seq},
        \texttt{Human},
        },
        ytick={0,...,1},
        ytick style={draw=none},
        enlargelimits=false,
        yticklabel style={xshift=2.5pt,font=\tt,scale=1.5},
        colorbar,
        colorbar style={
            ytick={0.00,0.20,0.40,0.60,0.80,1.00},
            yticklabels={0.00,0.20,0.40,0.60,0.80,1.00},
            yticklabel={\pgfmathprintnumber\tick},
            yticklabel style={font=\tt, scale=1.5,
            		/pgf/number format/fixed,
			/pgf/number format/fixed zerofill,
			/pgf/number format/precision=2}
        },
        point meta min=0.0,
        point meta max=1.0,
        nodes near coords={\pgfmathprintnumber\pgfplotspointmeta},
        nodes near coords black white/.style={
            small value/.style={
                yshift=-10pt,
                text=black,
                /pgf/number format/fixed,
                /pgf/number format/precision=3,
                /pgf/number format/zerofill=true,
                scale=1.5,
            },
            large value/.style={
                yshift=-10pt,
                text=white,
                /pgf/number format/fixed,
                /pgf/number format/precision=3,
                /pgf/number format/zerofill=true,
                scale=1.5,
            },
            every node near coord/.style={
                check for zero/.code={
                    \pgfmathfloatifflags{\pgfplotspointmeta}{0}{
                        \pgfkeys{/tikz/coordinate}
                    }{
                        \begingroup
                        \pgfkeys{/pgf/fpu}
                        \pgfmathparse{\pgfplotspointmeta<#1}
                        \global\let\result=\pgfmathresult
                        \endgroup
                        %
                        %
                        \pgfmathfloatcreate{1}{1.0}{0}
                        \let\ONE=\pgfmathresult
                        \ifx\result\ONE
                            \pgfkeysalso{/pgfplots/small value}
                        \else
                            \pgfkeysalso{/pgfplots/large value}
                        \fi
                    }
                },
                check for zero,
            },
        },
        nodes near coords black white=0.5,
    ]
        \addplot[
            matrix plot,
            mesh/cols=2,
            point meta=explicit,draw=gray
        ] table [meta=C] {
            x y C
0 0 0.967
1 0 0.033
0 1 0.015
1 1 0.985
         };
    \end{axis}
\end{tikzpicture}
\end{adjustbox}
&
\begin{adjustbox}{scale=0.875}
\begin{tikzpicture}[scale=0.5]
    \begin{axis}[
        width=7.5cm,
        height=7.5cm,
	colormap={bluewhite}{color=(white) rgb255=(100,149,237)},
        xticklabels={
        \texttt{Seq2Seq},
        \texttt{Human},
        },
        xtick={0,...,1},
        xtick style={draw=none},
	xticklabel style={anchor=east,rotate=60,yshift=-5pt,font=\tt,scale=1.5},
        yticklabels={
        \texttt{Seq2Seq},
        \texttt{Human},
        },
        ytick={0,...,1},
        ytick style={draw=none},
        enlargelimits=false,
        yticklabel style={xshift=2.5pt,font=\tt,scale=1.5},
        colorbar,
        colorbar style={
            ytick={0.00,0.20,0.40,0.60,0.80,1.00},
            yticklabels={0.00,0.20,0.40,0.60,0.80,1.00},
            yticklabel={\pgfmathprintnumber\tick},
            yticklabel style={font=\tt, scale=1.5,
            		/pgf/number format/fixed,
			/pgf/number format/fixed zerofill,
			/pgf/number format/precision=2}
        },
        point meta min=0.0,
        point meta max=1.0,
        nodes near coords={\pgfmathprintnumber\pgfplotspointmeta},
        nodes near coords black white/.style={
            small value/.style={
                yshift=-10pt,
                text=black,
                /pgf/number format/fixed,
                /pgf/number format/precision=3,
                /pgf/number format/zerofill=true,
                scale=1.5,
            },
            large value/.style={
                yshift=-10pt,
                text=white,
                /pgf/number format/fixed,
                /pgf/number format/precision=3,
                /pgf/number format/zerofill=true,
                scale=1.5,
            },
            every node near coord/.style={
                check for zero/.code={
                    \pgfmathfloatifflags{\pgfplotspointmeta}{0}{
                        \pgfkeys{/tikz/coordinate}
                    }{
                        \begingroup
                        \pgfkeys{/pgf/fpu}
                        \pgfmathparse{\pgfplotspointmeta<#1}
                        \global\let\result=\pgfmathresult
                        \endgroup
                        %
                        %
                        \pgfmathfloatcreate{1}{1.0}{0}
                        \let\ONE=\pgfmathresult
                        \ifx\result\ONE
                            \pgfkeysalso{/pgfplots/small value}
                        \else
                            \pgfkeysalso{/pgfplots/large value}
                        \fi
                    }
                },
                check for zero,
            },
        },
        nodes near coords black white=0.5,
    ]
        \addplot[
            matrix plot,
            mesh/cols=2,
            point meta=explicit,draw=gray
        ] table [meta=C] {
            x y C
0 0 0.961
1 0 0.039
0 1 0.038
1 1 0.962
         };
    \end{axis}
\end{tikzpicture}
\end{adjustbox}
\\[-1ex]
\adjustbox{scale=0.9}{(d) MLP}
&
\adjustbox{scale=0.9}{(e) DNN}
&
\adjustbox{scale=0.9}{(f) LSTM}
\end{tabular}
\caption{Confusion matrices after retraining (GloVe and T5-small)}\label{fig:conf_retrain}
\end{figure}

Figure~\ref{fig:conf_retrain} gives confusion matrices for all classification models in the 
case where models are retrained based on human text and text that was generated using GPT and then
transformed using T5-small, based on GloVe embeddings.
Recall that in this scenario, DNN is the best performing model, and from the confusion matrices
it is clear that this is due to the DNN having far fewer misclassifications of human text as 
Seq2Seq transformed GPT text.

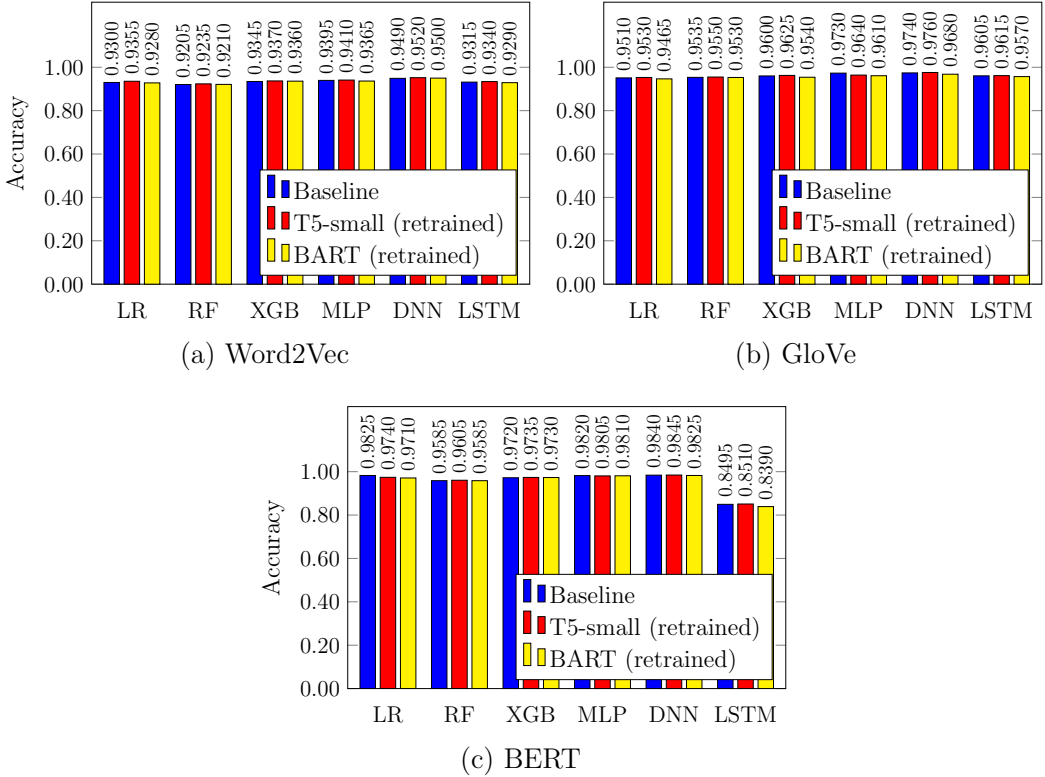
\begin{figure}[!htb]
    \centering\advance\tabcolsep by -2.5pt
    \begin{tabular}{cc}
    \begin{tikzpicture}[scale=0.9, every node/.style={scale=1.0}]
\pgfkeys{/pgf/number format/.cd,1000 sep={}}
\begin{axis}[
        width  = 0.52*\textwidth,
        height = 5.75cm,
        ymin=0.0,ymax=1.3,
        ytick={0.0, 0.2, 0.4, 0.6, 0.8, 1.0},
        major x tick style = transparent,
        ybar=5*\pgflinewidth,
        bar width=6.5pt,
        ylabel = {Accuracy},
        ylabel style = {scale=0.85},
        symbolic x coords={
        		LR, RF, XGB, MLP, DNN, LSTM
        },
        xticklabels={
        		LR, RF, XGB, MLP, DNN, LSTM
        },
	y tick label style={scale=0.80,
    		/pgf/number format/.cd,
   		fixed,
   		fixed zerofill,
    		precision=2},
        xtick = data,
        x tick label style={scale=0.8,
		},
        nodes near coords,
        every node near coord/.append style={rotate=90, scale=0.7,
        								   anchor=west, 
								   /pgf/number format/.cd,
								   fixed,
								   fixed zerofill,
								   precision=4},
        enlarge x limits=0.1125,
        legend cell align=left,
        legend pos=south east,
        legend style={nodes={scale=0.85},
        },
]
\addplot [fill=blue,opacity=1.00]
coordinates {
(LR, 0.9300)
(RF, 0.9205)
(XGB, 0.9345)
(MLP, 0.9395)
(DNN, 0.9490)
(LSTM, 0.9315)
};
\addlegendentry{Baseline}
\addplot [fill=red,opacity=1.00]
coordinates {
(LR, 0.9355)
(RF, 0.9235)
(XGB, 0.9370)
(MLP, 0.9410)
(DNN, 0.9520)
(LSTM, 0.9340)
};
\addlegendentry{T5-small (retrained)}
\addplot [fill=yellow,opacity=1.00]
coordinates {
(LR, 0.9280)
(RF, 0.9210)
(XGB, 0.9360)
(MLP, 0.9365)
(DNN, 0.9500)
(LSTM, 0.9290)
};
\addlegendentry{BART (retrained)}
\end{axis}
\end{tikzpicture}
    &
    \begin{tikzpicture}[scale=0.9, every node/.style={scale=1.0}]
\pgfkeys{/pgf/number format/.cd,1000 sep={}}
\begin{axis}[
        width  = 0.52*\textwidth,
        height = 5.75cm,
        ymin=0.0,ymax=1.3,
        ytick={0.0, 0.2, 0.4, 0.6, 0.8, 1.0},
        major x tick style = transparent,
        ybar=5*\pgflinewidth,
        bar width=6.5pt,
        symbolic x coords={
        		LR, RF, XGB, MLP, DNN, LSTM
        },
        xticklabels={
        		LR, RF, XGB, MLP, DNN, LSTM
        },
	y tick label style={scale=0.80,
    		/pgf/number format/.cd,
   		fixed,
   		fixed zerofill,
    		precision=2},
        xtick = data,
        x tick label style={scale=0.8,
		},
        nodes near coords,
        every node near coord/.append style={rotate=90, scale=0.7,
        								   anchor=west, 
								   /pgf/number format/.cd,
								   fixed,
								   fixed zerofill,
								   precision=4},
        enlarge x limits=0.1125,
        legend cell align=left,
        legend pos=south east,
        legend style={nodes={scale=0.85},
        },
]
\addplot [fill=blue,opacity=1.00]
coordinates {
(LR, 0.9510)
(RF, 0.9535)
(XGB, 0.9600)
(MLP, 0.9730)
(DNN, 0.9740)
(LSTM, 0.9605)
};
\addlegendentry{Baseline}
\addplot [fill=red,opacity=1.00]
coordinates {
(LR, 0.9530)
(RF, 0.9550)
(XGB, 0.9625)
(MLP, 0.9640)
(DNN, 0.9760)
(LSTM, 0.9615)
};
\addlegendentry{T5-small (retrained)}
\addplot [fill=yellow,opacity=1.00]
coordinates {
(LR, 0.9465)
(RF, 0.9530)
(XGB, 0.9540)
(MLP, 0.9610)
(DNN, 0.9680)
(LSTM, 0.9570)
};
\addlegendentry{BART (retrained)}
\end{axis}
\end{tikzpicture}
    \\
    \adjustbox{scale=0.9}{(a) Word2Vec}
    &
    \adjustbox{scale=0.9}{(b) GloVe}
    \\ \\[-1.25ex]
    \multicolumn{2}{c}{\begin{tikzpicture}[scale=0.9, every node/.style={scale=1.0}]
\pgfkeys{/pgf/number format/.cd,1000 sep={}}
\begin{axis}[
        width  = 0.52*\textwidth,
        height = 5.75cm,
        ymin=0.0,ymax=1.3,
        ytick={0.0, 0.2, 0.4, 0.6, 0.8, 1.0},
        major x tick style = transparent,
        ybar=5*\pgflinewidth,
        bar width=6.5pt,
        ylabel = {Accuracy},
        ylabel style = {scale=0.85},
        symbolic x coords={
        		LR, RF, XGB, MLP, DNN, LSTM
        },
        xticklabels={
        		LR, RF, XGB, MLP, DNN, LSTM
        },
	y tick label style={scale=0.80,
    		/pgf/number format/.cd,
   		fixed,
   		fixed zerofill,
    		precision=2},
        xtick = data,
        x tick label style={scale=0.8,
		},
        nodes near coords,
        every node near coord/.append style={rotate=90, scale=0.7,
        								   anchor=west, 
								   /pgf/number format/.cd,
								   fixed,
								   fixed zerofill,
								   precision=4},
        enlarge x limits=0.1125,
        legend cell align=left,
        legend pos=south east,
        legend style={nodes={scale=0.85},
        },
]
\addplot [fill=blue,opacity=1.00]
coordinates {
(LR, 0.9825)
(RF, 0.9585)
(XGB, 0.9720)
(MLP, 0.9820)
(DNN, 0.9840)
(LSTM, 0.8495)
};
\addlegendentry{Baseline}
\addplot [fill=red,opacity=1.00]
coordinates {
(LR, 0.9740)
(RF, 0.9605)
(XGB, 0.9735)
(MLP, 0.9805)
(DNN, 0.9845)
(LSTM, 0.8510)
};
\addlegendentry{T5-small (retrained)}
\addplot [fill=yellow,opacity=1.00]
coordinates {
(LR, 0.9710)
(RF, 0.9585)
(XGB, 0.9730)
(MLP, 0.9810)
(DNN, 0.9825)
(LSTM, 0.8390)
};
\addlegendentry{BART (retrained)}
\end{axis}
\end{tikzpicture}}
    \\
    \multicolumn{2}{c}{\adjustbox{scale=0.9}{(c) BERT}}
    \end{tabular}
    \caption{Accuracy comparison for retraining}
    \label{fig:accuracy-comp-retrain}
\end{figure}

For ease of comparison, in Figure~\ref{fig:accuracy-comp-retrain} we provide bar graphs 
comparing the results obtained for the baseline case of 
distinguishing GPT-generated text from human-generated
text (see Section~\ref{sect:case1}, above) to the
results in this section, which were obtained by retraining the models
on Seq2Seq-transformed text. We observe that the results
are virtually identical, indicating that retrained classification models can 
distinguish our transformed text from human text as easily
as they can distinguish the original GPT-generated text from human-generated text.

\subsection{Discussion}

Our experiments demonstrate that by 
post-processing GPT-generated text using a Seq2Seq approach, we can significantly 
degrade the performance of classification models that have been trained to
distinguish GPT-generated text from human-generated text. 
This indicates that Seq2Seq models can be useful 
in an adversarial learning scenario, where the goal is to defeat
machine learning and deep learning 
models that have been trained to distinguish AI-generated text from human-generated text.

Interestingly, we also found that by retraining classification models on text that had
been transformed by our Seq2Seq approach, we were able to distinguish Seq2Seq-transformed
text from human-generated text with high accuracy. In fact, the accuracies in this case were
essentially the same as those obtained for the baseline problem of distinguishing GPT-generated
text from human-generated text. This result indicates that although our Seq2Seq 
models degrade the performance of GPT-generated text classifiers, the transformed 
text can still be distinguished from human-generated text. That is, our Seq2Seq-transformed 
text is not sufficiently human-like to evade detection by classification models that are trained 
to distinguish between these types of data. This is somewhat surprising, since the Seq2Seq 
models were designed to ``humanize'' the GPT data.

\section{Conclusion and Future Work}\label{chap:conclusion}

Through a set of detailed experiments, we demonstrated the effectiveness of using the
fine-tuned sequence-to-sequence (Seq2Seq) models T5-small and BART to transform 
GPT-generated text into a less-easily detectable form. The evaluation across different
embeddings (Word2Vec, GloVe, and BERT) confirmed that all classifiers tested
(Logistic Regression, Random Forest, XGBoost,
Multilayer Perceptron, Deep Neural Network, 
and Long Short-Term Memory network)
experienced a significant drop in performance when tested on transformed GPT-generated text.
This result highlights the viability of Seq2Seq models in such an adversarial scenario.
However, learning models trained specifically to detect our Seq2Seq-transformed 
text performed well, indicating that such transformed text can still be distinguished from human-generated text.
Overall, these results validate our adversarial transformation pipeline,
while also clearly demonstrating that effective defense strategies are available.
In particular, this work demonstrates the importance of adversarial-style data 
augmentation for building more robust AI-generated text detection systems.

There remains considerable scope for future research. 
Improving the human-like qualities of Seq2Seq-transformed text is perhaps the most
obvious future research topic. While our Seq2Seq-transformed text 
degrades the performance of trained classifiers, 
it was insufficiently human-like to
evade retrained classifiers. In this context, 
other Seq2Seq models, such as Flan-T5~\cite{10.5555/3722577.3722647}, 
Mistral-7B~\cite{jiang2023mistral7b}, or LLaMA 2-13B~\cite{touvron2023llama2openfoundation},
might prove more effective.
Another, related area of future work would be to
incorporate adversarial fine-tuning loops~\cite{rafid2023adversarialtrainingusingfeedback}, 
where feedback from classifiers is used to guide the text transformation process. 
By implementing such a process, we can potentially develop adversarial models 
that actively learn how to reduce detectability.

\bibliographystyle{plain}
\bibliography{references}

\section*{Appendix}\label{app:a}

\titleformat{\section}{\normalfont\large\bfseries}{}{0em}{#1\ \thesection}
\setcounter{section}{0}
\renewcommand{\thesection}{\Alph{section}}
\renewcommand{\thesubsection}{A.\arabic{subsection}}
\setcounter{table}{0}
\renewcommand{\thetable}{A.\arabic{table}}
\setcounter{figure}{0}
\renewcommand{\thefigure}{A.\arabic{figure}}

This appendix provides representative examples showing the original human-written text, the 
GPT-generated text, and the corresponding transformed outputs generated by 
the Seq2Seq models T5-small and BART.

\subsection*{\large Example 1}

\noindent\textbf{Human:} Even if you are giving in to other people and doing nice things for them all of the time, you may not always have positive feelings towards them.You might feel resent towards people for taking up your time and energy. However, you might not be expressing these feelings and they may be building up in the form of resentments, or pent up anger towards other people. Try to express these feelings to help you see when you have been too accommodating to someone. For example, you might say something like, “I wish I had not given up my evening plans to watch my sister’s children. I really needed that time to relax and I feel angry at her for asking me to give that up.” The next time someone asks you to do something that you don’t really want to do, consider whether it might lead to resentment. If the answer is yes, then you might want to say no. 

\bigskip

\noindent\textbf{ChatGPT:} In the Considering Your Thoughts and Feelings section, it is crucial to note any lingering feelings of resentment that may arise. As a person who is often too nice for their own good, it is not uncommon to suppress negative emotions and prioritize the needs of others above your own. However, resentment can quietly build up over time if you consistently neglect your own thoughts and feelings. Take a moment to reflect on situations where you may have felt taken advantage of or overlooked. Acknowledging and addressing these feelings of resentment is essential for maintaining a healthy balance in your relationships and ensuring your own well-being.

\bigskip

\noindent\textbf{Transformed T5-small:} In the Considering Your Thoughts and Feelings section, it’s important to recognize any lingering feelings of resentment that might come up. If you are someone who tends to be too nice, it is common to push aside negative emotions and put others needs before your own. But over time, resentment can slowly build if you keep ignoring your own thoughts and feelings. Take a moment to think about situations where you may have felt taken advantage of or overlooked. Recognizing and working through these feelings is essential for keeping a healthy balance in your relationships and protecting your own well being.

\bigskip

\noindent\textbf{Transformed BART:} In the Considering Your Thoughts and Feelings section, it is important to notice any lingering resentment that may surface. Being someone who is often too nice can lead to suppressing negative emotions and putting the needs of others ahead of your own. However, resentment can quietly grow if you continue to overlook your own feelings. Reflect on times when you may have felt taken advantage of or ignored. Acknowledging these feelings is important for maintaining healthy relationships and looking after your own well being.

\subsection*{\large Example 2}

\noindent\textbf{Human:} Throw away the ink in your vessel, the needle from the gun, the gloves, and the rest of the equipment that you used. This stuff cannot be reused, if you want to give yourself sterile, clean, and effective tattoos. Only use new, clean products when you're giving a tattoo.

\bigskip

\noindent\textbf{ChatGPT:} Clean your workspace thoroughly before attempting to give yourself a tattoo to ensure a hygienic and safe environment for the procedure. Begin by removing any unnecessary clutter and organizing your materials, such as tattoo needles, ink, and gloves, in a clean and sterile manner. Disinfect all surfaces with an appropriate cleaning solution, paying close attention to the area where you will be working. Additionally, ensure that you have access to running water and soap to maintain proper hand hygiene throughout the process. A clean workspace is vital for preventing infections and promoting the healing process post-tattooing.

\bigskip

\noindent\textbf{Transformed T5-small:} Before giving yourself a tattoo, make sure to thoroughly clean your workspace to create a hygienic and safe environment. Start by clearing away any unnecessary clutter and organizing your materials, such as tattoo needles, ink, and gloves, in a clean and sterile setup. Disinfect all surfaces with a proper cleaning solution, focusing especially on the area where you plan to work. Also, ensure you have easy access to running water and soap to maintain good hand hygiene during the process. Keeping your workspace clean is crucial for preventing infections and helping the tattoo heal properly afterward.

\bigskip

\noindent\textbf{Transformed BART:} Thoroughly clean your workspace before attempting to give yourself a tattoo to ensure a hygienic and safe setting. Remove any unnecessary clutter and organize your materials, including tattoo needles, ink, and gloves, in a clean and sterile way. Disinfect all surfaces using an appropriate cleaning solution, paying special attention to the area you will be using. Make sure you have running water and soap available to maintain hand hygiene throughout. A clean workspace is essential for preventing infections and supporting the healing process after tattooing.

\subsection*{\large Example 3}

\noindent\textbf{Human:} Some medications, like blood thinners or aspirin, can be detrimental to healing and actually increase bleeding and swelling. There are also some natural remedies and supplements that can be concerning. As long as you disclose everything you are taking to your doctor, she can evaluate the safety of your medication regimen and make sure that nothing conflicts with wound healing or worsens bleeding. The doctor will tell you what medications need to be stopped prior to the surgery, when they should be stopped, and when to restart. Also, other medications may need to be continued straight through the surgery.

\bigskip

\noindent\textbf{ChatGPT:} When it comes to healing after plastic surgery, it is crucial to consult your doctor regarding any medications or supplements you routinely use. This information is essential for the subsequent medical steps involved in the healing process. By informing your doctor about your regular intake of medications or supplements, they can provide accurate guidance on whether to continue, modify, or temporarily halt their usage to ensure a smooth recovery. It is important to prioritize open communication with your doctor to optimize your healing journey.

\bigskip

\noindent\textbf{Transformed T5-small:} When healing after plastic surgery, it is important to talk to your doctor about any medications or supplements you regularly take. Sharing this information is essential for the next steps in the healing process. By letting your doctor know about your regular intake, they can guide you on whether you should continue, adjust, or temporarily stop certain medications to support a smooth recovery. Open communication with your doctor is key to making sure your healing journey goes as well as possible.

\bigskip

\noindent\textbf{Transformed BART:} During the healing process after plastic surgery, it is important to consult your doctor about any medications or supplements you use regularly. This information is critical for determining the next medical steps in your recovery. By informing your doctor, they can accurately advise whether you should continue, adjust, or temporarily stop using certain medications to promote better healing. Prioritizing clear communication with your doctor is essential for optimizing your recovery.

\end{document}